\pdfoutput=1

\documentclass[11pt]{article}

\usepackage[final]{acl}

\usepackage{times}
\usepackage{latexsym}

\usepackage[T1]{fontenc}

\usepackage[utf8]{inputenc}
\usepackage{amssymb}

\usepackage{microtype}

\usepackage{inconsolata}

\usepackage{graphicx}

\usepackage{algorithm}
\usepackage{algpseudocode}
\usepackage{xcolor}
\usepackage{amsmath}
\usepackage{booktabs}
\usepackage{multirow}
\usepackage{makecell}
\usepackage{pifont}
\usepackage{siunitx}
\usepackage{float}
\usepackage{caption}
\usepackage{cleveref}
\usepackage{fancyhdr}

\fancypagestyle{firstpagefooter}{
    \fancyhf{} 
    
    \fancyfoot[C]{
        \small 
        \textit{Proceedings of the 2025 Conference on Empirical Methods in Natural Language Processing}, pages 13041–13068 \\
        November 4-9, 2025 \copyright 2025 Association for Computational Linguistics
    }
}

\newcommand{\Ours}{CFGM}

\definecolor{mygreen}{RGB}{43, 138, 62}
\definecolor{myred}{RGB}{201, 42, 42}

\newcommand*\colourcheck[1]{%
  \expandafter\newcommand\csname #1check\endcsname{\textcolor{#1}{\ding{51}}}%
}
\colourcheck{mygreen}

\newcommand*\colourmark[1]{%
  \expandafter\newcommand\csname #1mark\endcsname{\textcolor{#1}{\ding{55}}}%
}
\colourmark{myred}

\title{Coarse-to-Fine Grounded Memory for LLM Agent Planning}

\vspace{-4mm}
\author{Wei Yang\textsuperscript{1}, Jinwei Xiao\textsuperscript{1}, Hongming Zhang\textsuperscript{1}$^\dagger$, Qingyang Zhang\textsuperscript{1}, Yanna Wang\textsuperscript{1}, Bo Xu\textsuperscript{1}$^\dagger$\\
\textsuperscript{1}National Key Laboratory of Cognition and Decision Intelligence for Complex Systems, \\ Institution of Automation, Chinese Academy of Sciences \\
\texttt{\{yangwei2023,xiaojinwei2024,hongming.zhang\}@ia.ac.cn}\\
\texttt{\{zhangqingyang2019,wangyanna2013,boxu\}@ia.ac.cn}
\\
\small{
   $^\dagger$\textbf{Correspondence:} \href{mailto:hongming.zhang@ia.ac.cn}{hongming.zhang@ia.ac.cn}, \href{mailto:boxu@ia.ac.cn}{boxu@ia.ac.cn}
 }
}

\begin{document}
\maketitle
\begin{abstract}

Recent advancements in Large Language Models (LLMs) have driven growing interest in LLM-based agents for complex planning tasks. To avoid costly agent training, many studies adopted memory mechanism that enhances LLM with offline experiences or online trajectory analysis. However, existing works focus on single-granularity memory derived from dynamic environmental interactions, which are inherently constrained by the quality of the collected experiences. This limitation, in turn, constrain the diversity of knowledge and the flexibility of planning. We propose Coarse-to-Fine Grounded Memory (\Ours{}), a novel framework that grounds coarse-to-fine memories with LLM, thereby fully leverage them for flexible adaptation to diverse scenarios. \Ours{} grounds environmental information into coarse-grained focus points to guide experience collection in training tasks, followed by grounding of actionable hybrid-grained tips from each experience. At inference, \Ours{} retrieves task-relevant experiences and tips to support planning. When facing environmental anomalies, the LLM grounds the current situation into fine-grained key information, enabling flexible self-QA reflection and plan correction.
Extensive experiments on AlfWorld, Webshop and ScienceWorld demonstrate that \Ours{} significantly outperforms competitive baselines and comprehensively optimizes memory-enhanced LLM Agent system.

\end{abstract}

\section{Introduction}\label{sec:intro}

\begin{figure}[ht]
  \includegraphics[width=\linewidth]{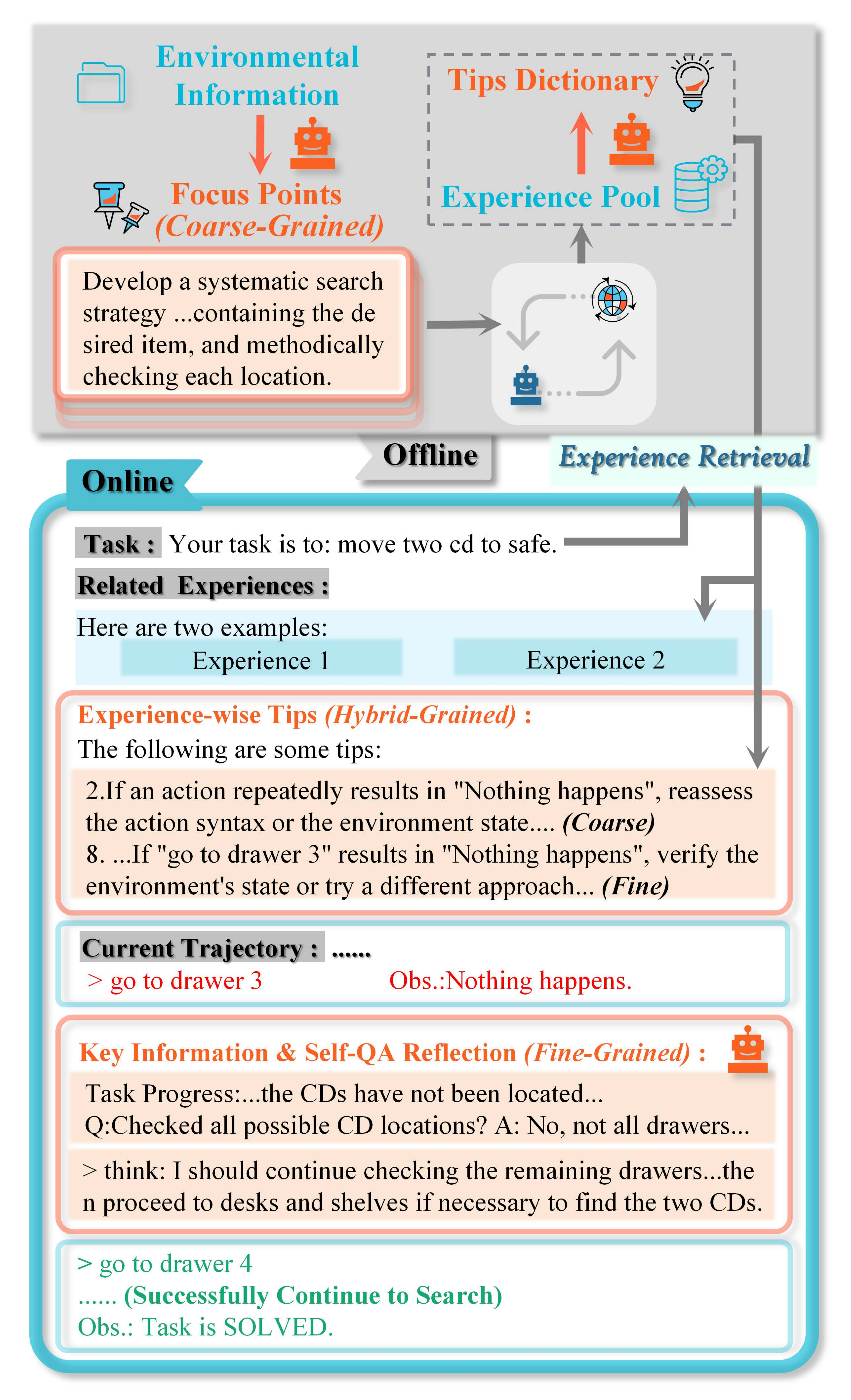} 
  \caption {Conceptual overview of \Ours{}. During the offline training, our method first extracts coarse-grained focus points to guide experience collection, then distills hybrid-grained tips from these experiences. At the online inference time, it retrieves relevant experiences and tips for planning. When encountering anomalies, the system identifies fine-grained key details for adaptive self-QA and plan adjustment.}
  \label{fig:concept}
\end{figure}
\thispagestyle{firstpagefooter}

The remarkable progress of LLMs has ignited intense interest in their deployment as autonomous agents capable of tackling complex, long-horizon tasks within interactive environments~\citep{Wang2023, Xi2023, Gur2024, Wang2024}. These LLM-based agents leverage extensive world knowledge~\citep{Zhao2023} and reasoning capabilities to formulate policies~\citep{yao2023react}. However, significant challenges hinder their practical effectiveness. Agents often suffer from inefficient exploration, yielding sparse or irrelevant data~\citep{Arumugam2025,zhang2025beta,zhang2025edge}. They also struggle with reasoning derailment in partially observable environments, where incomplete observations cause cascading errors~\citep{kim-etal-2024-qube}. Existing methods attempt to mitigate these issues using memory mechanisms, such as storing offline experiences and insights~\citep{Zhao2024}, extracting context-aware guidelines~\citep{Fu2024}, or checking belief states with fixed-question reflection~\citep{kim-etal-2024-qube}.


However, recent studies depend on single-granularity memory derived from dynamic environmental interactions, whose effectiveness is limited by the quality of the collected experiences. This constraint consequently reduces the variety of knowledge and limits the adaptability of planning. These limitations lead to explicit outcomes manifested in three aspects: (1) When confronted with challenging planning tasks, the initial exploration efficiency of LLMs is often suboptimal, constraining the quality of memory. (2) Current approaches extract knowledge either at a holistic level or a scenario-specific level, both of which lead to homogenized granularity in the acquired knowledge. (3) Although directly using manually designed questions to reflect on trajectory information in short-term memory can yield decent results, it lacks generalizability across environments.

To bridge this gap, we introduce \textbf{C}oarse-to-\textbf{F}ine \textbf{G}rounded \textbf{M}emory (\Ours{}), a novel agent framework that leverages the internal knowledge of LLM to guide the memory collection and utilization, thereby enhancing their planning capabilities in complex environments. As illustrated in Figure \ref{fig:concept}, \Ours{} operates through three progressively memory grounding stages to ensure that the agent’s planning and actions remain continuously anchored in LLM's prior understanding of memory. Guided by environmental information (initial memory), the LLM first identifies coarse-grained focus points for task planning. This preliminary understanding enables the collection of higher-quality experiences. For each one of the offline experiences (long-term memory), LLM grounds the knowledge it contains into high-level principles and detailed techniques, thereby constructing an experience-wise hybrid-grained tips dictionary. During online interactions, the most relevant successful experiences and tips are retrieved to enhance LLM planning as context. Upon facing environmental anomalies, the agent conducts a fine‑grained key information analysis of the current state (short-term memory) , followed by a self-QA reflection grounded in both past successes and the present situation to effectively correct planning. By grounding explicit coarse-to-fine memories with LLM’s internal knowledge, \Ours{} enables the LLM to autonomously consider diverse information in complex planning environments and optimizes the collection and utilization of memory based on its reasoning outcomes. Because of this, \Ours{} is able to enhance the quality of collected memories, diversify experiential knowledge, and provide more flexible error‑correction support during planning. Finally, this enables LLMs to tackle complex tasks more effectively, even preserving compatibility with powerful closed‑source models. We evaluate \Ours{} on three diverse interactive planning benchmarks. Our results demonstrate that \Ours{} significantly outperforms existing memory-enhanced agent systems, showcasing improved performance and robustness.

Our three key contributions are as follows: (i) We propose Coarse-to-Fine Grounded Memory (\Ours{}), a novel framework that enhances LLM agents by systematically grounding memory with LLM's internal knowledge during experiences collection, tips extraction and adaptive planning. (ii) We validate the effectiveness of \Ours{} across multiple challenging benchmarks, achieving state-of-the-art performance compared to strong memory-enhanced LLM agent system baselines. (iii) We perform comprehensive ablation studies and analyses on the framework's core components, which show coarse-to-fine grounded memory enhances the quality of collected memory experiences, constructs hybrid-grained tips with more useful information in offline phase, and enables flexible and effective reflection during online planning.

\section{Related Work}


\subsection{LLM-based Agents}

As an emerging technology, LLM-based agents employ LLM which provides language understanding and generation capabilities as the core of planning~\citep{Yue2025}. Compared to traditional AI agents, LLM-based agents demonstrate higher levels of autonomy and reasoning capabilities~\citep{Luo2025}. LLM-based agents have been applied in various domains, including software engineering~\citep{Jin2024}, question answering~\citep{Yue2025}, scientific research~\citep{Xiao2024}, web navigation~\citep{Deng2023}, mathematical reasoning~\citep{ahn-etal-2024-large}, embodied intelligence~\citep{zhao-etal-2024-epo}, and search and recommendation systems~\citep{zhang2025survey}. In this paper, we aim to construct high-performance agents targeting diverse planning task with coarse-to-fine grounded memory.

\subsection{Experience-based Optimization for Agents}
Experience-based optimization focuses on improving the performance of agents by leveraging historical data and accumulated knowledge~\citep{Li2024g,Li2024c,Zhao2024,bai2024bayesian,Chen2024c,Fu2024,qian-etal-2024-experiential,fu-etal-2024-msi,zhangamulet}. This optimization provides a mechanism for LLM agents to learn and adapt over time without adjusting the model's parameters~\citep{Du2025,Zheng2025,song-etal-2024-trial}. ExpeL~\citep{Zhao2024} exemplifies this approach through prompt-based learning with experience recall and insight extraction. AutoGuide~\citep{Fu2024} extracts and applies context-aware guidelines from experiences. However, these methods often struggle with inefficient exploration in complex planning tasks due to limited prior task understanding, leading to suboptimal memory collection. Furthermore, when extracting knowledge from memories, they typically rely on single-grained information, restricting the agent's holistic comprehension. To overcome these issues, \Ours{} leverages LLM-generated focus points to guide exploration toward critical dynamics, reducing redundancy. Additionally, it extracts hybrid-grained tips from experiences to enhance LLMs' comprehensive memory understanding.

\subsection{Feedback-based Optimization for Agents}

Feedback-based optimization involves using various types of online feedback to guide the learning and improvement of agents. This feedback can come from external models~\citep{Bo2025,Huang2023,Yao2023,zhang2024exploiting,bai2024efficient,sun2025retrieval}, or the agent itself~\citep{huang-etal-2024-queryagent, wang-etal-2024-boosting-llm,Liang2024c,zhang-etal-2024-agent,Zhou2024a,Shinn2023,cui2025self}. Self-reflection and self-correction enable agents to iteratively improve their outputs by identifying and addressing errors~\citep{Luo2025}. Reflexion~\citep{Shinn2023} preforms trial-and-error attempts on online inference trajectories, reflects on the mistakes, and applies the lessons learned to subsequent attempts. QuBE ~\citep{kim-etal-2024-qube} conducts belief-state reflection with fixed-question templates, which addresses errors in tasks that allow only a single attempt. To enable more flexible short-term memory reflection in single-attempt planning compared to existing approaches, \Ours{} employs adaptive Self-QA reflection that dynamically leverage key information extracted from online trajectory information with LLM.

\section{Method}

\begin{figure*}[ht]
  \includegraphics[width=\linewidth]{figures/framework_917.pdf} 
  \caption {Framework of \Ours{}. \Ours{} collect experiences offline with coarse-grained focus points grounded from environmental information, followed by extracting the hybrid-grained tips grounded from trajectories of each experience to construct tips dictionary. Then, agent's online planning will be enhanced by retrieved experiences and tips, in which the fine-grained self-QA reflection is activated by the key information grounded from current situation and relevant history when anomaly observed. The \textcolor[RGB]{251,97,32}{orange} arrow represents the memory grounding process.}
  \label{fig:framework}
\end{figure*}

To tackle the current challenges that memory-augmented agents encounter during planning, we introduce an LLM‑based agent system characterized by a coarse‑to‑fine grounding of the model’s internal knowledge into explicit memories. This system leverages inherent LLM knowledge to enhance the quality of collected memories, diversify high‑level experiential knowledge, and provide more flexible error‑correction support during planning, ultimately enabling LLMs to handle complex decision‑making tasks more effectively. The framework is depicted in Figure \ref{fig:framework} and comprises three components: coarse‑grained focus‑driven experience collection, hybrid‑grained experience‑wise tips extraction and fine‑grained key information adaptive planning. The prompt template related to memory grounding in the entire framework are presented in Appendix \ref{sec:prompt_templates}.

\begin{algorithm}[th]
\footnotesize  
\caption{Coarse‑Grained Focus‑Driven Experience Collection.}


\begin{algorithmic}[1]
\State $N = len(\mathcal{T_\text{train}}),n = 0$
\State $FP = \text{LLM}_\text{Focus}(Desc_\text{env}, F_\text{manual})$ 
\While{task $n < N$}
    \State $t_n \gets \mathcal{T_\text{train}}[n],\nu_{n,0} \gets \text{“”} $
    \For{trial $z = 0$ to $Z$}
        \State Initialize Trajectory  $\tau_{n,z} \gets \text{env.reset($t_n$)}$ 
        \For{timestep $ i=0 $ to  $ H $ }
            \State $ a_i \gets \text{LLM}_\text{ReAct}(a_i\mid \tau_{n,z}, F_\text{manual},\nu_{n,z},FP) $
            \State $ o_{i+1}, \texttt{done} \gets \text{env.step}(a_i) $
            \State $ \tau_{n,z} \gets \tau_{n,z} \cup \{(a_i,o_{i+1})\} $
            \If{$ \texttt{done} $}
                \State break
            \EndIf
        \EndFor
        \State $ \mathcal{B} \gets \mathcal{B} \cup (t_n,\tau_{n,z}) $
        \If{ $ \texttt{done} $ or $z = Z$ }
            \State $ n \gets n+1 $
            \State break
        \Else
            \State $ \nu_{n,z+1} \gets \text{concat}(\nu_{n,z} + \text{LLM}_\text{Reflect}(\tau_{n,z})) $
        \EndIf
    \EndFor
\EndWhile

\State \Return $ \mathcal{B} $

\end{algorithmic}
\label{alg:exp_gather}
\end{algorithm}

\subsection{Coarse‑Grained Focus‑Driven Experience Collection}

To collect usable and diverse experiential memories, we adopted a synergistic approach Reflexion~\cite{Shinn2023} for collaborative thinking and retry-reflection to collect diverse experiences. Additionally, we enabled the LLM to ground environmental foundational information to obtain focus points that guide LLM with their own preliminary understanding of the environment to collect higher-quality experience. The workflow of this component is illustrated in the pseudo-code of \Cref{alg:exp_gather}. Specifically, at the initiation of experience collection, the agent $\text{LLM}_\text{Focus}$ conducts a comprehensive analysis of the environmental description $Desc_\text{env}$ and manual trajectories few-shot $F_\text{manual}$ using its internal knowledge. Through coarse-grained examination of task-related foundational information, the agent can generate some preliminary thoughts for handling the environmental tasks. Guided by these focus points $FP$ on training task sets $\mathcal{T_\text{train}}$, the agent $\text{LLM}_\text{ReAct}$ executes up to $Z + 1$ rounds of trial attempts involving think-act iterations. During this process, failures trigger introspective reflection $\text{LLM}_\text{Reflect}(\tau_{n,z})$, with updates incorporated into task-specific reflections $\nu_{n,z+1}$ to assist subsequent attempts until task $t_n$ completion. High-quality trajectory information $\tau_{n,z}$ collected under focus points guidance is stored in the experience pool $\mathcal{B}$ with the corresponding task $t_n$. Both successful and failed trajectories in success task will subsequently support tip generation and key information reflection phases.

\begin{algorithm}[th]
\footnotesize  
\caption{Hybrid‑Grained Experience‑Wise Tips Extraction.}


\begin{algorithmic}[1]
\State $N = len(\mathcal{T_\text{train}})$
\State Construct fail/success tuples of the same tasks in  $\mathcal{B}$ :
\State \( C_\text{compare} = \{t_1:(\tau_1^\text{success},\tau_{1,0}^\text{fail}, ...), ..., \)
\State \(\quad \quad \quad \quad
t_i:(\tau_i^\text{success},\tau_{i,0}^\text{fail}, ...), ...\}\)

\State Construct success trajectories in  $\mathcal{B}$ :

\State \( C_\text{success} = \{t_1:\tau_1^\text{success}, ...,t_i:\tau_i^\text{success}, ...\}\)

\For{task \( m = 0 \) to \( N - 1 \)}
    \State $t_n \gets \mathcal{T_\text{train}}[n]$
    \If{$ t_n \in  C_\text{compare} $}
        \State $TD[t_n] = \text{LLM}_\text{Tips}(C_\text{compare}[t_n])$
        \State $TD[t_n] \gets (TD[t_n],$
        \State $\quad \quad \quad \quad
        \text{LLM}_\text{Tips}(C_\text{success}[t_n],TD[t_n]))$
        \State continue
    \EndIf

    \If{$ t_n \in  C_\text{success} $}
        \State $TD[t_n] = \text{LLM}_\text{Tips}(C_\text{success}[t_n]))$
    \EndIf
\EndFor

\State \Return $ TD $

\end{algorithmic}
\label{alg:tips_extraction}
\end{algorithm}

\subsection{Hybrid‑Grained Experience‑Wise Tips Extraction}
Because the experience collection stage follows a repeated trial-and-error paradigm~\citep{sutton2018reinforcement,dong2020deep}, many tasks stored in the experience pool are accompanied by both successful and failed trajectories. These trajectories contain not only fine‑grained scene-specific details but also more high-level coarse‑grained insights that are applicable across similar tasks. We refer to such hybrid‑grained knowledge as tips. To help the LLM make better use of this hybrid-grained information, the current module leverages the model’s internal knowledge offline to convert the multiple trajectories of a given task experience into a set of tips that blend hybrid-grained perspectives. 

The complete workflow of this module is illustrated in \Cref{alg:tips_extraction}. We begin by re-indexing all trajectory data in the experience pool $\mathcal{B}$ according to their associated task $t_n$. Trajectories that contain both successful and failed attempts are aggregated into the compare experience dictionary $C_\text{compare}$, while those consisting a successful attempt are placed in the success experience dictionary $C_\text{success}$. Next, we iterate over every training task $t_n$. If $t_n$ appears in the $C_\text{compare}$, we prompt the $\text{LLM}_\text{Tips}$ to compare the task’s successful and failed trajectories and, drawing on its internal knowledge, extract tips that help prevent the observed errors. We then ask the LLM to focus exclusively on successful trajectories to refine additional success-specific factors to expand tips. The resulting tips are stored under the corresponding task $t_n$ in the Tips Dictionary $TD$. For tasks that the LLM can solve directly in a successful attempt, we perform the same reflective analysis on their successful trajectories, aiming to derive a moderate number of tips that remain applicable in similar scenarios.


\subsection{Fine‑Grained Trajectory Information Adaptive planning}

After completing the offline collection of experiences and extraction of relevant hybrid-grained knowledge, we need to dynamically utilize these auxiliary information online during evaluation based on fine-grained current status information. In addition to leveraging auxiliary information, the agent model inevitably encounters unexpected scenarios during evaluation and planning. To mitigate the resulting planning derailment, we employ LLM to perform adaptive grounding of corrective planning by analyzing the fine-grained current trajectory status and analogous successful cases.

The detailed workflow of the aforementioned evaluation reasoning is presented in \Cref{alg:adapt_reason}. For task $t_m$ from evaluation set $\mathcal{T_\text{evaluation}}$, we conduct a prior analysis based on task description retrieval before inference. Using the Faiss library~\cite{johnson2019billion}, we retrieve the top-k most similar experiences $E_\text{sim}$ from the Experience Pool $\mathcal{B}$ with textual embedder $\mathcal{E}$. These experiences' successful trajectory contents $\tau_{h}^\text{success}$ and corresponding tips knowledge $TD[t_h]$ are integrated into Similar Trajectories content $ST$ and relevant Experience Tips $ET$ , which serve as contextual assistance for the decision-making of policy model $\text{LLM}_\text{\Ours{}}$ during inference. During the reasoning process for task $t_m$, we implement real-time monitoring of observed information $o_{i+1}$ and trigger alerts for specific abnormal observations. When triggered, we employ the Key Information Extraction model $\text{LLM}_\text{KIE}$ to perform a structured, fine-grained analysis of the current trajectory content $\tau_{m}$, thereby obtaining grounded current trajectory content as key information $KI$. Subsequently, by synthesizing $KI$, $\tau_{m}$, and $ST$ as contextual elements, we further instruct the Key Information Reflection model $\text{LLM}_\text{KIR}$ to conduct self-reflective question-answer $qa_i$, ultimately deriving a final corrective plan $ref_i$. The $ref_i$ is then integrated into the current trajectory content to guide $\text{LLM}_\text{\Ours{}}$ in implementing adaptive responses to exceptional situations. 

\begin{algorithm}[th]
\footnotesize  
\caption{Fine‑Grained Trajectory Information Adaptive Planning.}

\begin{algorithmic}[1]
\State $M = len(\mathcal{T_\text{evaluation}})$
\For{task \( m = 1 \) to \( M \)}
    \State $t_m \gets \mathcal{T_\text{evaluation}}[m] $
    \State Initialize trajectory \( \tau_{m} \leftarrow \text{env.reset($t_m$)} \)
    \State \( E_\text{sim} \leftarrow \texttt{Faiss}(t_m, \mathcal{B}, \mathcal{E}, k) \)
    \State $ET \gets \text{“”}, ST \gets \text{“”}$
    \For{experience task \( (t_h,\tau_{h}^\text{success})\) in \( E_\text{similar tasks} \)}
        \State $ET \gets \text{concat}(ET+TD[t_h])$
        \State $ST \gets \text{concat}(ST+\tau_{h}^\text{success})$
    \EndFor
    \For{timestep \( i = 1\) to \( H \)}
        \State \( a_i \leftarrow \text{LLM}_\text{\Ours{}}(a_i\mid \tau_{m}, ET, ST) \)
        \State \( o_{i+1}, \texttt{done} \leftarrow \text{env.step}(a_i) \)
        \State current step $s_i \gets (a_i,o_{i+1})$
        \If{$trigger(o_{i+1})$}
            \State $KI \gets \text{LLM}_\text{KIE}(\tau_{m})$
            \State $(qa_i,ref_i) \gets \text{LLM}_\text{KIR}(\tau_{m}, KI, ST)$
            \State $s_i \gets \text{concat}(s_i+ref_i)$
        \EndIf
        \State \( \tau_{m} \leftarrow \tau_{m} \cup \{s_i\} \)
        \If{\( \texttt{done} \)}
            \State break
        \EndIf
    \EndFor
\EndFor
\end{algorithmic}
\label{alg:adapt_reason}
\end{algorithm}

\section{Experiments}

\subsection{Experimental Settings}
\label{sec:experimental_settings}

We evaluate \Ours{} on three challenging interactive environment to assess its effectiveness and capabilities, particularly how its coarse-to-fine grounding mechanism enhance LLM agent performance. We refer Appendix \ref{sec:environment_details} for detailed information of main environments and WebArena~\citep{zhou2023webarena} used in ablation study.

\noindent \textbf{Environments.} The basic introduction of environment is as follows: (i) AlfWorld is an virtual household agent benchmark where the agent performs household tasks based on natural language instructions. It requires navigation and object manipulation~\citep{Shridhar2020}. (ii) WebShop is an interactive environment simulating online shopping, where the agent needs to find and purchase a product matching specific criteria by interacting with a website~\citep{Yao2022}. (iii) ScienceWorld is a text-based environment focusing on elementary science tasks, requiring complex reasoning and interaction with the simulated world~\citep{wang-etal-2022-scienceworld}.


\noindent \textbf{Evaluation.} 
Following standard practices in interactive agent evaluation, we use Success Rate (SR) as the primary metric across all environments. Specifically, SR measures completing the task within time limits for AlfWorld, purchasing the item that matches all attributes for WebShop, and successful task completion for ScienceWorld. For WebShop, we also report the mean reward, which quantifies the quality of the purchased product based on the matching of required attributes. 



\noindent \textbf{Baselines.} We compare \Ours{} against some mainstream LLM agent methods, most of which are memory-augmented agents: (i) ReAct is a foundational method that interleaves reasoning steps with actions in the environment~\citep{yao2023react}. (ii) ExpeL learns from past experiences by extracting insights and retrieving relevant trajectories~\citep{Zhao2024}. (iii) AutoGuide generates context-aware guidelines from offline experiences to guide the agent at test time~\citep{Fu2024}. (iiii) QuBE enhances agent reasoning by constructing belief state reflection with human-curated questions to alleviate reasoning derailment~\citep{kim-etal-2024-qube}.


\noindent \textbf{Implementation Details.} Unless otherwise specified (e.g., in model generalization studies), we use GPT-4-Turbo as the agent model and GPT-4o as the knowledge extraction model (like insight extraction model in ExpeL) for all baselines and \Ours{}. Additional implementation details are presented in Appendix \ref{sec:implementation_details}.

\subsection{Main Results}


\begin{table*}[htbp] 
    \centering

    \tabcolsep=0.1cm 

    \begin{tabular}{l ccc c cc c} 
        \toprule
        \multirow{2}{*}[-10pt]{\textbf{Method}} 
        & \multicolumn{3}{c}{\textbf{Features}} 
        & \multirow{2}{*}[6.825pt]{\textbf{AlfWorld}} 
        & \multicolumn{2}{c}{\textbf{WebShop}} 
        & \multirow{2}{*}[6.825pt]{\textbf{ScienceWorld}}\\ 
        \cmidrule(lr){2-4} 
        \cmidrule(lr){5-5} 
        \cmidrule(lr){6-7} 
        \cmidrule(lr){8-8} 
        & \makecell{Off.\\Exp.} & \makecell{Traj.\\Ref.} & \makecell{Gro.\\Mem.} 
        & \makecell{Success Rate (SR)$\uparrow$} & \makecell{Reward$\uparrow$} & \makecell{SR$\uparrow$} & \makecell{SR$\uparrow$}\\
        \midrule 

        ReAct & \myredmark & \myredmark & \myredmark & 80.60\% $\pm$ 0.68\% & 58.6 $\pm$ 1.0 & 37\% $\pm$ 2\% & 43\% $\pm$ 1\%\\
        ExpeL & \mygreencheck & \myredmark & \myredmark & 81.34\% $\pm$ 0.85\% & 62.2 $\pm$ 1.3 & 42\% $\pm$ 3\% & 57\% $\pm$ 2\% \\
        AutoGuide & \mygreencheck & \myredmark & \myredmark & 83.58\% $\pm$ 0.77\% & 73.3 $\pm$ 1.4 & 47\% $\pm$ 2\% & N/A \\
        QuBE & \myredmark & \mygreencheck & \myredmark & 84.33\% $\pm$ 0.73\%  & N/A & N/A & N/A \\
        ExpeL + QuBE & \mygreencheck & \mygreencheck & \myredmark & \underline{85.07\% $\pm$ 1.03\%} & \underline{75.6 $\pm$ 1.3} & \underline{49\% $\pm$ 3\%} & \underline{65\% $\pm$ 2\%} \\
        \Ours{} & \mygreencheck & \mygreencheck & \mygreencheck & \bfseries 91.00\% $\pm$ 0.82\% & \bfseries 85.0 $\pm$ 1.3 & \bfseries 57\% $\pm$ 3\% & \bfseries 74\% $\pm$ 2\% \\
        \bottomrule 

    \end{tabular}

    \caption{Main results across AlfWorld, WebShop, and ScienceWorld benchmarks. Off. Exp. refers to collecting offline experiences, Traj. Ref. refers to reflection based on trajectories during online inference, and Gro. Mem. refers to leveraging grounded memory with LLM's knowledge. Best in bold, second-best underlined. \Ours{} consistently outperforms the baselines on all domains, highlighting the importance of comprehensive memory grounding and the scalability of \Ours{} in diverse environments.} 
    \label{tab:main_results} 
\end{table*}


Table \ref{tab:main_results} presents the main performance comparison of \Ours{} (Ours) against the baselines for the three benchmarks. Based on these results, we conduct separate analyses focusing on the framework's effectiveness and environmental generalizability.


\textbf{Effectiveness of \Ours{}.} As shown in Table \ref{tab:main_results}, \Ours{} outperforms ReAct across all environments: AlfWorld SR (+10.40\%), WebShop SR (+20\%), and ScienceWorld SR (+31\%). This performance boost is due to \Ours{}'s coarse-to-fine grounding approach, which actively integrates LLM knowledge into structured memories~\citep{zhang2023replay}. Compared to other memory-augmented agent baselines, our method also achieves superior task performance. This improvement stems from two key aspects: firstly, our approach leverages both offline experiences from long-term memory to assist online planning and incorporates short-term memory reflection on current trajectories; secondly, we utilize the internal knowledge of LLMs to ground the memories, enabling the agent to better comprehend and utilize past experiences. Additionally, \Ours{} consistently and significantly outperforms the ExpeL + QuBE baseline. This demonstrates \Ours{}'s superiority is not merely an artifact of combining two orthogonal techniques.Furthermore, we wish to re-emphasize that CFGM is not a simple combination of existing ideas. Its components are deeply integrated and synergistic with coarse-to-fine grounded memory. Specifically, for example, the coarse-grained Focus Points are not just an independent pre-processing step; they fundamentally improve the quality of the memories collected. This, in turn, enhances the effectiveness of both the hybrid-grained Experience-wise Tips and the fine-grained Key Information Reflection. This synergistic "1+1>2" effect is a core contribution of our work and distinguishes CFGM from a simple integrated pipeline of unrelated modules.

\textbf{Robustness in  Diverse Environments.} \Ours{} shows strong scalability and robustness, achieving significant gains in diverse environments. Specifically, it reaches 91.00\% SR in AlfWorld, 0.85 Reward and 57\% SR in WebShop, and 74\% SR in ScienceWorld. These tasks range from long action sequences planning to dynamic web navigation and scientific experiments, where agents typically struggle with planning derailment and inefficient exploration. These results demonstrate the generalization capability of \Ours{}'s memory grounding approach across different environments. Detailed \Ours{}'s experimental trajectory examples for each environment are provided in the Appendix~\ref{sec:examples} with ReAct's comparative trajectories.

\subsection{Ablation Studies}

After evaluating the performance of the framework, we conduct extensive experiments to assess the framework's effectiveness and the design of memory grounding process of each component.


\noindent \textbf{Effectiveness of Different Components.} Fistly, we evaluate the impact of each core component of \Ours{}—coarse-grained focus points (FP), hybrid-grained experience tips (ET), and fine-grained key information reflection (KIR)—by adding them individually or in combination to a ReAct baseline on AlfWorld. As shown in Table \ref{tab:module-ablation}, adding coarse-to-fine grounded memory individually provides consistent gains over the ReAct baseline. This demonstrates that grounded memory based on the LLM's internal knowledge enhances the agent's planning capability across all granularity levels. By integrating all three components, \Ours{} achieves the highest SR of 91.00\%, demonstrating the enhancement effects of grounded memory at different granularities on LLM planning are cumulative and validating the rationality of our coarse-to-fine framework design. It is worth noting that removing FP causes SR to drop to 88.06 \%, showing FP enhances the positive effects of ET  and KIR on LLM planning performance by improving the quality of collected memories.

\begin{table}[htbp]
  \centering
  \setlength{\tabcolsep}{3pt}
  \begin{tabular}{l|cccccc}
    \toprule
    \textbf{FP}  & \myredmark     & \mygreencheck & \myredmark     & \myredmark     & \myredmark     & \mygreencheck \\
    \textbf{ET}  & \myredmark     & \myredmark    & \mygreencheck  & \myredmark     & \mygreencheck  & \mygreencheck \\
    \textbf{KIR} & \myredmark     & \myredmark    & \myredmark     & \mygreencheck  & \mygreencheck  & \mygreencheck \\
    \midrule
    \textbf{SR\%}  & 80.60          & 85.82         & 86.57          & 85.82          & 88.06          & 91.00         \\
    \bottomrule
  \end{tabular}
  \caption{Module ablation experiment results on AlfWorld. FP refers to using focus points to guide experience collection, ET refers to extracting experience-wise tips, and KIR refers to question-answer self-reflection based on key information during online planning.}
  \label{tab:module-ablation}
\end{table}

\begin{figure}[htbp]
  \centering
  \includegraphics[width=0.95\columnwidth]{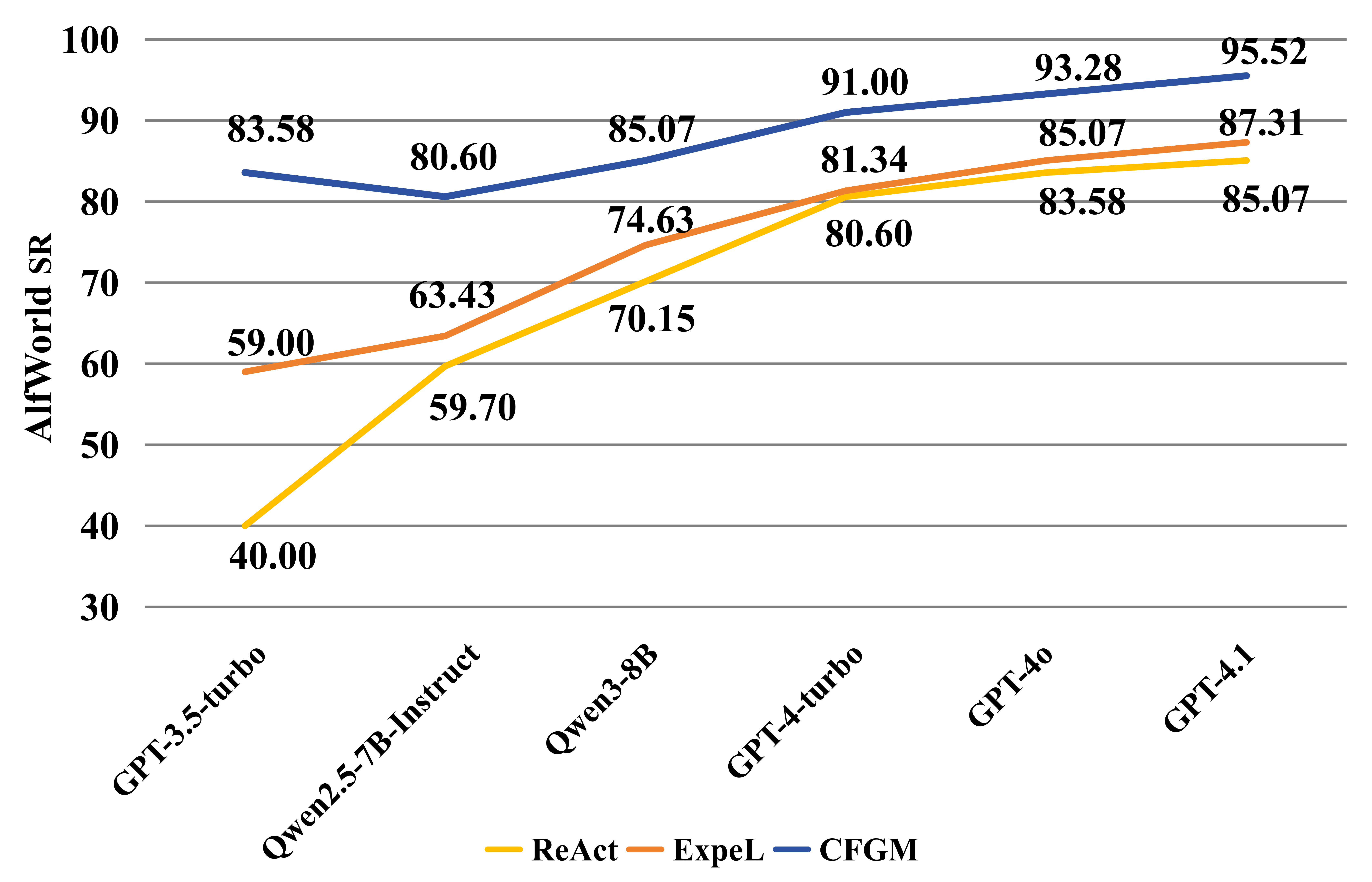} 
  \caption {The SR achieved by different methods using various models on AlfWorld. \Ours{} demonstrates strong generalization across different models and consistently outperforms the baselines.}
  \label{fig:models}
\end{figure}

\noindent \textbf{Generalizability across Different Models.} Having examined the detailed mechanisms of individual component, we proceed to evaluate the framework's overall generalizability across diverse agent models. For closed-source GPT-based models up to GPT-4o, we choose GPT-4o as knowledge extraction model. For other models (Qwen, GPT-4.1), the component rels on the model itself, allowing evaluation of the framework's self-sufficiency with different base LLMs. As presented in Figure \ref{fig:models}, from smaller open-source ones (Qwen2.5-7B-Instruct) to powerful closed-source ones (GPT-4.1), \Ours{} consistently outperforms both ReAct and ExpeL by a significant margin, which proves that the grounded memory in \Ours{} exhibits universal effectiveness in enhancing planning capabilities across different LLM models. Having analyzed the effectiveness and generalizability of grounded memory, we further explore the design and capability of grounded memory at key component.

\noindent \textbf{Impact of Experience Retrieval Scope on Tips Quality.} For the hybrid-grained tips used online, we investigate how the number of retrieved experiences (top-$k$) affects the tips' quality. Specifically, we evaluate the impact by the SR on AlfWorld using a ReAct + ET setup. As shown in Table \ref{tab:topk_ablation}, the baseline ReAct ($k=0$) has an SR of 80.60\%. Introducing just one relevant experience's tips ($k=1$) improves SR to 82.09\%. Performance peaks at $k=2$ and $k=3$, both achieving 86.57\% SR. Increasing k further to 5 leads to a drop in performance (81.34\%), due to the inclusion of less relevant or potentially conflicting hybrid-grained tips. This suggests that retrieving a moderate, focused set of the most relevant experiences ($k=2$ or $3$) provides the highest quality hybrid-grained tips to guide the LLM agent.


\begin{table}[H]
  \centering
  \setlength{\tabcolsep}{2.5pt} 
  \begin{tabular}{c|ccccc}
    \toprule  
    \textbf{ $k$ }           & 0     & 1     & 2     & 3     & 5     \\
    \midrule
    \textbf{ SR }  & 80.60\% & 82.09\% & 86.57\% & 86.57\% & 81.34\% \\
    \bottomrule 
  \end{tabular}
  \caption{The impact of different Top-$k$ experience retrieval scope on the quality of tips on AlfWorld. The retrieved top-2 experiences yield tip sets with optimal quality and retrieval efficiency.}
  \label{tab:topk_ablation}
\end{table}

\noindent \textbf{Experience Tips' Generalizability to Out-of-Distribution Environments.} To test the out-of-domain capability of tips across different domains but relevant tasks, we conduct a transfer learning experiment from WebShop to WebArena-Shopping that is a distinct domain but contain relevant tasks. We extract hybrid-grained tips from WebShop experiences and apply them to WebArena-Shopping. For the domain adaptation, we align the tips to WebArena-Shopping based on GPT-4-Turbo. As presented in Table \ref{tab:transfer_learning}, the transferred tips achieves 25.1\% SR, significantly outperforming all the baselines, which demonstrates that the hybrid-grained tips grounded from experiences can capture generalizable knowledge, enabling effective transfer to related but distinct out-of-domain environments.


\begin{table}[H]
  \centering
  \begin{tabular}{lc}
    \hline
    \textbf{Method} & \textbf{WebArena–Shopping SR} \\
    \hline
     ReAct & 10.2\% $\pm$ 0.5\%\\
     ExpeL & 18.5\% $\pm$ 0.9\%\\
     AutoGuide & 20.4\% $\pm$ 0.7\%\\
     \Ours{} & 25.1\% $\pm$ 0.8\%\\
    \hline
  \end{tabular}
  \caption{Out-of-distribution generalization of different method's extracted experience knowledge from WebShop on the 98 WebArena–Shopping tasks. hybrid-grained tips of \Ours{} perform superior out-of-domain generalization compared to existing knowledge extraction agents after transfer.}
  \label{tab:transfer_learning}
\end{table}

\noindent \textbf{Comparative Analysis of Online Reflection.} To explore the rationality of KIR's memory grounding design, we analyze the effectiveness of different online reflection mechanisms on AlfWorld. Table \ref{tab:kir-ablation} compares KIR against two other reflection variants:  a template‑free Self-QA reflection without key information and a QA reflection via fixed questioning template like QuBE~\citep{kim-etal-2024-qube}), both applied on top of a ReAct agent. As shown in Table \ref{tab:kir-ablation}, KIR demonstrates superior corrective assistance for agent planning compared to QA Reflection (Fixed Question), which indicates that fine-grained grounded memory can achieve more effective reflection outcomes with greater flexibility than manually human-crafted question-answering reflection approaches. The performance improvement of KIR over Self-QA Reflection further demonstrates that prior analysis of short-term memory can significantly enhance subsequent comprehension and reasoning effectiveness when utilizing grounded memory.


\begin{table}[htbp]
  \centering
  \setlength{\tabcolsep}{4pt} 
  \begin{tabular}{lc}
    \hline
    \textbf{Method}     & \textbf{AlfWorld SR} \\
    \hline
    Self-QA Reflection          & 84.33\%              \\
    QA Reflection(Fixed Question)  & 84.33\%             \\
    Key Information Reflection  & 85.82\%              \\
    \hline
  \end{tabular}
  \caption{The comparative analysis of different reflection. Fine-grained grounded memory enables more flexible and effective reflection than manual QA approaches. Prior short-term memory analysis (key information) further enhances comprehension and reasoning when leveraging grounded memory.}
  \label{tab:kir-ablation}
\end{table} 

\noindent \textbf{The Overall Efficiency of \Ours{}.} To assess the practical efficiency of \Ours{} compared against the baseline methods, we analyzed the token consumption and interaction turns of the framework on the AlfWorld benchmark. As shown in Table \ref{tab:consum}, during offline processing, \Ours{}'s token cost is comparable to ExpeL and notably more efficient than AutoGuide, which show the efficient offline traininng process of \Ours{}. Additionally, in the online phase, \Ours{} is the most efficient framework among the memory-augmented baselines in terms of both tokens per step and average interaction turns. This reduction in steps can lead to lower overall latency and cost, demonstrating that the comprehensive memory grounding leads to more efficient, less redundant agent behaviors.

\begin{table}[H]
  \centering
  \setlength{\tabcolsep}{2pt}
  \begin{tabular}{lccc}
    \hline
    \textbf{Method} & \textbf{Off. Tokens} & \textbf{On. Tokens} & \textbf{On. Turns}\\
    \hline
     ReAct & - & 1734.6 & 19.01\\
     ExpeL & 3888.2 & 5125.5 & 17.32\\
     AutoGuide & 4873.4 & 4809.4 & 16.55\\
     QuBE & - & 4752.3 & 17.21\\
     \Ours{} & 4068.5 & 4681.4 & 14.32\\
    \hline
  \end{tabular}
  \caption{Average token consumption for processing a single training data sample during the offline phase on AlfWorld (Off. Tokens). Average token consumption per step and average interaction turns per complete trajectory during online inference on AlfWorld (On. Tokens and On. Turns).}
  \label{tab:consum}
\end{table}
\section{Conclusion}

In this paper, we propose an LLM agent system enhanced with coarse-to-fine grounded memory. \Ours{} relies on the LLM’s intrinsic knowledge to analyze coarse-grained environmental information, hybrid-grained experiences' trajectories, and fine-grained current trajectory. This process yields knowledge that enhances the LLM’s comprehension and interactive performance in complex planning scenarios. Experimental results across three challenging benchmarks demonstrate the superiority of \Ours{} over existing memory-enhanced LLM agent systems. Through comprehensive optimization of the agent’s framework, \Ours{} obtains higher-quality memory, more comprehensive experiential knowledge, and more flexible online reflection compared to other baselines. Additional experiments further validate the rationality and effectiveness of each component of \Ours{}. These findings offer valuable insights into the full utilization of memory knowledge in LLM agent systems.

\section*{Limitations}

As a memory-augmented LLM Agent system, our work focuses primarily on memory optimization utilization through LLM internal knowledge grounding, yet still presents several limitations: 1) Although the collection efficiency for training tasks has improved, it remains challenging to generate sufficient effective experiences when confronted with extremely small training sets. Therefore, how to effectively expand limited training task sets constitutes an important research direction for future investigation. 2) While leveraging experiences from similar tasks during online inference demonstrates strong auxiliary effects under moderate task similarity, their effectiveness may diminish if the trajectory of experience is very long, potentially introducing redundancy or interference in LLM planning. Future work can explore dynamic filtering mechanisms to adaptively retain the most relevant planning information from the long trajectory, further enhancing the robustness of the approach across diverse scenarios.

\section*{Acknowledgments}

This work is supported by the National Key R\&D Program of China (No.2022ZD0116405).

\bibliography{anthology, custom}

%

\clearpage
\appendix
\section*{Appendix}
\label{sec:appendix}

\section{Environment Details}
\label{sec:environment_details}
\subsection{AlfWorld~\citep{Shridhar2020}}
AlfWorld is a simulated environment that bridges the gap between language-based reasoning and embodied action by aligning tasks in a text-based interactive fiction environment with corresponding tasks in a visually rich, embodied 3D simulator. This unique structure allows for the study of agents capable of both high-level abstract planning and low-level grounded execution.

AlfWorld encompasses a diverse set of household tasks, requiring agents to perform a sequence of actions to achieve a specified goal. These tasks are categorized into six main types:

\begin{itemize}
    \item Pick \& Place (e.g., “move a plate to the coffee table”) - the agent must find an object of the desired type, pick it up, find the correct location to place it, and move it there.
    \item Examine in Light (e.g., “examine a book under the lamp”) - the agent must find an object of the desired type, locate and turn on a light source with the desired object in-hand.
    \item Clean \& Place (e.g., “clean the knife and move it to the drawer”) - the agent must find an object of the desired type, pick it up, go to a sink or a basin, wash the object by turning on the faucet, then find the correct location to place it, and move it there.
    \item Heat \& Place (e.g., “heat a mug and move it to the coffee table”) - the agent must find an object of the desired type, pick it up, go to a microwave, heat the object turning on the microwave, then find the correct location to place it, and move it there.
    \item Cool \& Place (e.g., “move a cool bottle to the countertop”) - the agent must find an object of the desired type, pick it up, go to a fridge, put the object inside the fridge and cool it, then find the correct location to place it, and move it there.
    \item Pick Two \& Place (e.g., “move two pencils to the drawer”) - the agent must find an object of the desired type, pick it up, find the correct location to place it, move it there, then look for another object of the desired type, pick it up, return to previous location, and move it there with the other object.
\end{itemize}


\begin{table*}[htp]
\centering
\setlength{\tabcolsep}{5mm}
\begin{tabular}{c c}
\toprule
\textbf{Retrieval Parameters} &  \\ 
\midrule
Vectorstore & Faiss \\
Retriever Type & kNN \\
Embedder & all-mpnet-base-v2 \\
Experience Retrieval Scope & 2 \\ 
Top-k Experience's Tips Selection  & 2 \\
\midrule
\textbf{Agent Parameters} &  \\
\midrule
Max Reflection Retries & 3 \\
Focus Points Generation Model & gpt-4-turbo-2024-04-09 \\
Policy Model & gpt-4-turbo-2024-04-09 \\
Reflection Model & gpt-4-turbo-2024-04-09 \\
Tips Generation Model & gpt-4o-2024-08-06 \\
Key Information Extraction Model & gpt-4-turbo-2024-04-09 \\
Key Information Reflection Model & gpt-4-turbo-2024-04-09 \\
Max Tips from Compare Traj. & 5 \\
Max Tips from Success Traj. & 3 \\
\midrule
\textbf{AlfWorld-specific Parameters} &  \\ 
\midrule
Max Number of Environment Steps & 20 \\ 
Number of Experiment Tasks & 134 \\ 
\midrule
\textbf{WebShop-specific Parameters} &  \\ 
\midrule
Max Number of Environment Steps & 15 \\ 
Number of Experiment Tasks & 134 \\ 
\midrule
\textbf{ScienceWorld-specific Parameters} &  \\
\midrule
Max Number of Environment Steps & 80 \\ 
Number of Experiment Tasks & 100 \\ 
\midrule
\textbf{WebArena-Shopping-specific Parameters} &  \\
\midrule
Max Number of Environment Steps & 20 \\ 
Number of Experiment Tasks & 90 \\ 
Tips Alignment Model & gpt-4-turbo-2024-04-09 \\
\bottomrule
\end{tabular}
\vspace{5pt}
\caption{Retrieval, Agent and Environment Parameters.}
\label{tab:parameters}
\end{table*}

\subsection{WebShop~\cite{Yao2022}}

WebShop is an interactive environment that simulates a real-world e-commerce website, designed as a benchmark for evaluating interactive planning and grounded language understanding in agents. The environment features a substantial catalog of 1.18 million real-world products, each with various attributes and options. Agents are tasked with completing shopping missions based on 12,087 diverse, crowd-sourced natural language instructions. The WebShop benchmark also includes over 1,600 human demonstration trajectories, which can be utilized for validating the task and supporting imitation learning.

The core task for an agent in WebShop is to navigate the simulated website, search for products, identify items that match the criteria given in the instruction, potentially customize options (e.g., color, size), and finally purchase the correct item. The agent interacts with the environment by issuing a variety of actions, such as typing search queries, clicking on product links or buttons, selecting product options, and executing a purchase command. Success is measured by whether the purchased item exactly matches the instruction’s requirements, yielding a binary success rate (SR).

A task episode concludes when the agent issues a purchase action or a maximum step limit is reached. The agent receives a reward based on the degree to which the purchased item matches the specified attributes and options in the initial instruction, including price constraints. The reward function is designed to reflect the quality of the purchased item against the user's requirements. Specifically, the reward \(r\) is calculated as:

{\small
\begin{align}
r = \frac{|U_{\text{att}} \cap Y_{\text{att}}| + |U_{\text{opt}} \cap Y_{\text{opt}}| + \mathbb{I}[y_{\text{price}} \le u_{\text{price}}]}{|U_{\text{att}}| + |U_{\text{opt}}| + 1} \cdot r_{\text{type}}
\end{align}
}
where $U_{\text{att}}$ and $Y_{\text{att}}$ are the sets of attribute names from the user query and the purchased product, respectively; $U_{\text{opt}}$ and $Y_{\text{opt}}$ are the sets of optional attribute values; $\mathbb{I}[y_{\text{price}} \le u_{\text{price}}]$ is an indicator function which is 1 if the purchased product's price is less than or equal to the user's preferred price, and 0 otherwise. $r_{\text{type}}$ is a term related to textual matching between the selected product's title and the target product's title, defined as:

{\small
\begin{align}
r_{\text{type}} =
\begin{cases}
0, & \text{if TextMatch} = 0 \\
0.1, & \text{if TextMatch} < 0.1 \\
0.5, & \text{if TextMatch} \le 0.2 \\
1, & \text{otherwise}.
\end{cases}
\end{align}
}
The term "TextMatch" denotes the textual overlap of pronouns, nouns, and proper nouns between the selected product's title and the target product's title.

\subsection{ScienceWorld~\citep{wang-etal-2022-scienceworld}}
ScienceWorld is a text-based virtual environment designed to evaluate an agent's scientific reasoning and procedural task completion abilities, benchmarked at the level of a standard elementary school science curriculum.

ScienceWorld features a rich, dynamic, and interactive simulated world. This includes multiple interconnected locations (e.g., kitchen, workshop, laboratory, greenhouse), populated with over 200 common household and science-related objects. These objects possess various properties (e.g., temperature, conductivity, state of matter) and afford a wide range of actions. The underlying simulation engines model simplified thermodynamics, electrical circuits, chemical reactions, and biological processes, allowing for complex interactions and state changes within the environment.

The benchmark comprises 30 diverse task types, spanning 10 distinct topics from the elementary science curriculum. Examples include tasks related to changing states of matter (e.g., melting ice, boiling water), understanding life cycles (e.g., growing a plant), exploring basic physics (e.g., building simple electrical circuits), and conducting simple chemistry experiments. For each task type, multiple variations are procedurally generated, altering object properties, initial locations, and specific sub-goals to test for generalization. Agents interact with the environment by issuing text commands from a predefined set of action templates (e.g., "open door", "take apple from table", "use thermometer on beaker"), and receive textual observations describing the current state of their surroundings and the outcomes of their actions. The average optimal decision depth for these tasks can be significant, often requiring a long sequence of correct actions to achieve the goal.

\subsection{WebArena-Shopping \citep{zhou2023webarena}}
WebArena is a suite of realistic and reproducible web environments designed for the development and evaluation of autonomous agents. The platform features fully operational, self-hostable web applications that emulate real-world websites across various domains.The WebArena–Shopping environment is implemented as part of the WebArena suite, providing a realistic and reproducible e‑commerce setting for interactive planning agents.

\section{More Implementation Details}
\label{sec:implementation_details}
To fairly compare with ExpeL, we employ four-fold validation for all experiments. We train on one half of the dataset and evaluate on the other half, and vice versa. All results include the mean and standard error of the results across the folds. The parameter information for the experiments conducted in each environment is shown in Table \ref{tab:parameters}.

We adopted different experimental task selection strategies for different environments. In the case of ALFWorld~\citep{Shridhar2020}, we utilized the 134 solvable tasks that ReAct and ExpeL used, which comprised of 6 different task types in a virtual household environment: Pick (24), Clean (18), Heat (31), Cool (23), Examine (21), and Pick two (17). Similarly, for WebShop~\cite{Yao2022}, we evaluated using the same 100 tasks used by ReAct and ExpeL. For ScienceWorld~\citep{wang-etal-2022-scienceworld}, we assess performance using 100 tasks sampled from test set, which comprised of diverse task types with environmental steps up to 80: task-10-use-thermometer (10), task-2-power-component (5), task-2a-test-conductivity(10), task-3-find-animal (10), task-3-find-living-thing (10), task-3-find-non-living-thing (10), task-3-find-plant (10), task-4-grow (8), task-5-chemistry (11), task-6-lifespan (3), task-7-identify-life-stage (9), task-1-change-the-state-of-matter (4).

\section{Prompt Templates}
\label{sec:prompt_templates}
\subsection{Focus Points Generation Model}
\begin{figure*}[htbp]
  \centering
  \includegraphics[width=\linewidth]{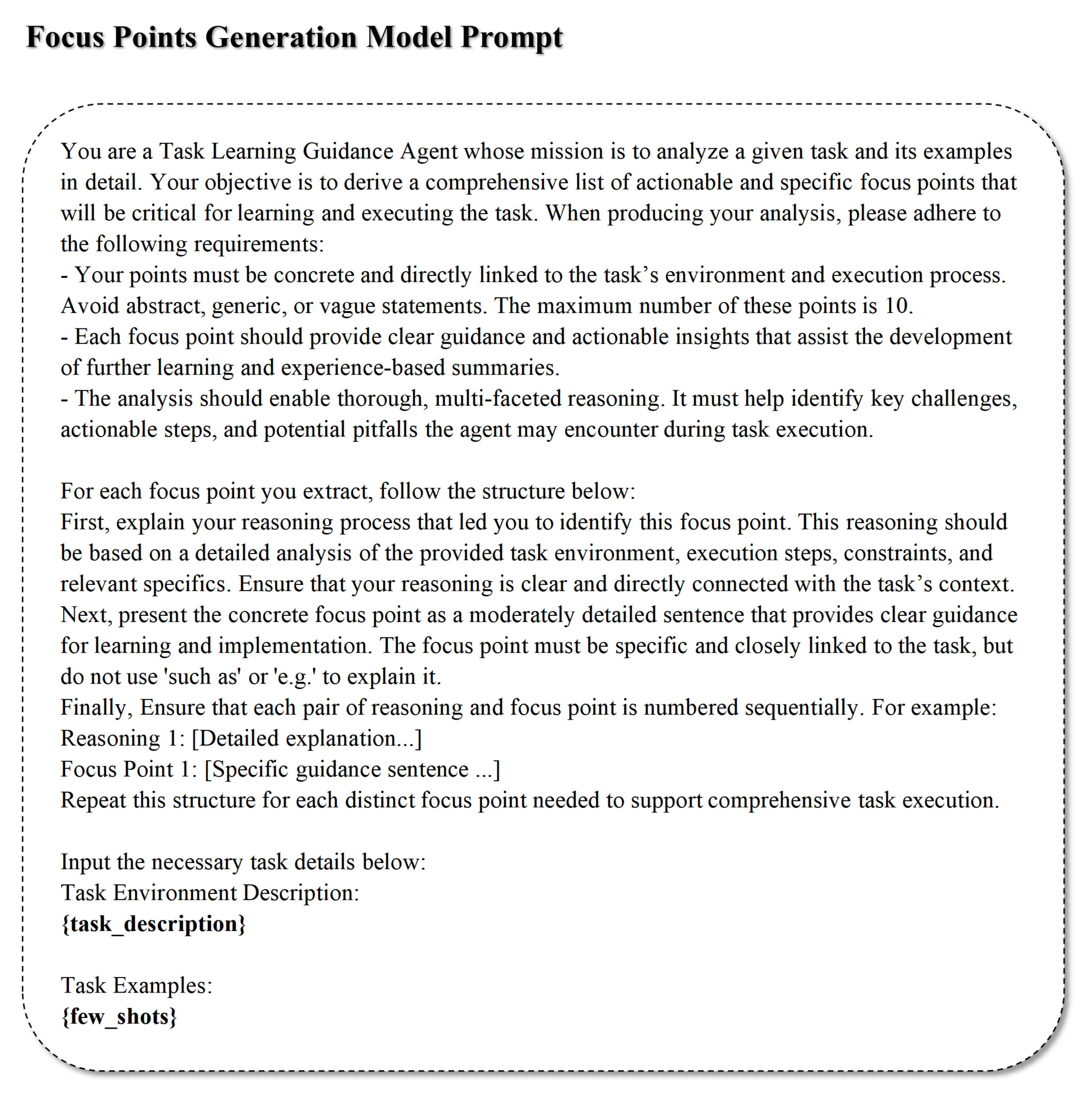} 
  \caption {The prompt template of focus points generation model in the ALFWorld, WebShop and ScienceWorld.}
  \label{fig:fp_prompt}
\end{figure*}

We present our prompt template of focus points generation model in the ALFWorld, WebShop and ScienceWorld in Figure \ref{fig:fp_prompt}.

\subsection{Experience-Wise Tips Generation Model}
\begin{figure*}[htbp]
  \centering
  \includegraphics[width=\linewidth]{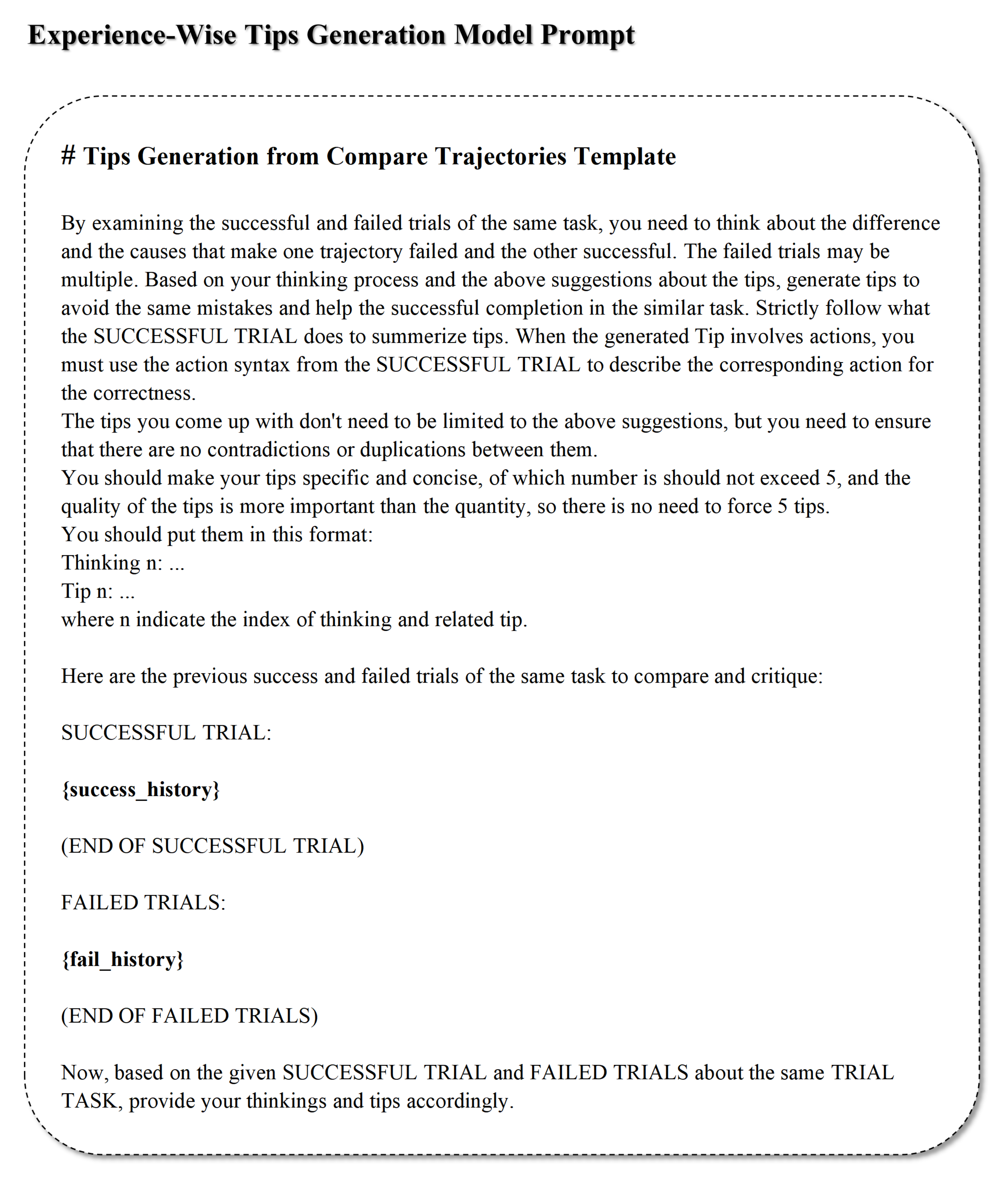} 
  \caption {The prompt template of tips generation from compare trajectories.}
  \label{fig:tips_compare}
\end{figure*}

\begin{figure*}[htbp]
  \centering
  \includegraphics[width=\linewidth]{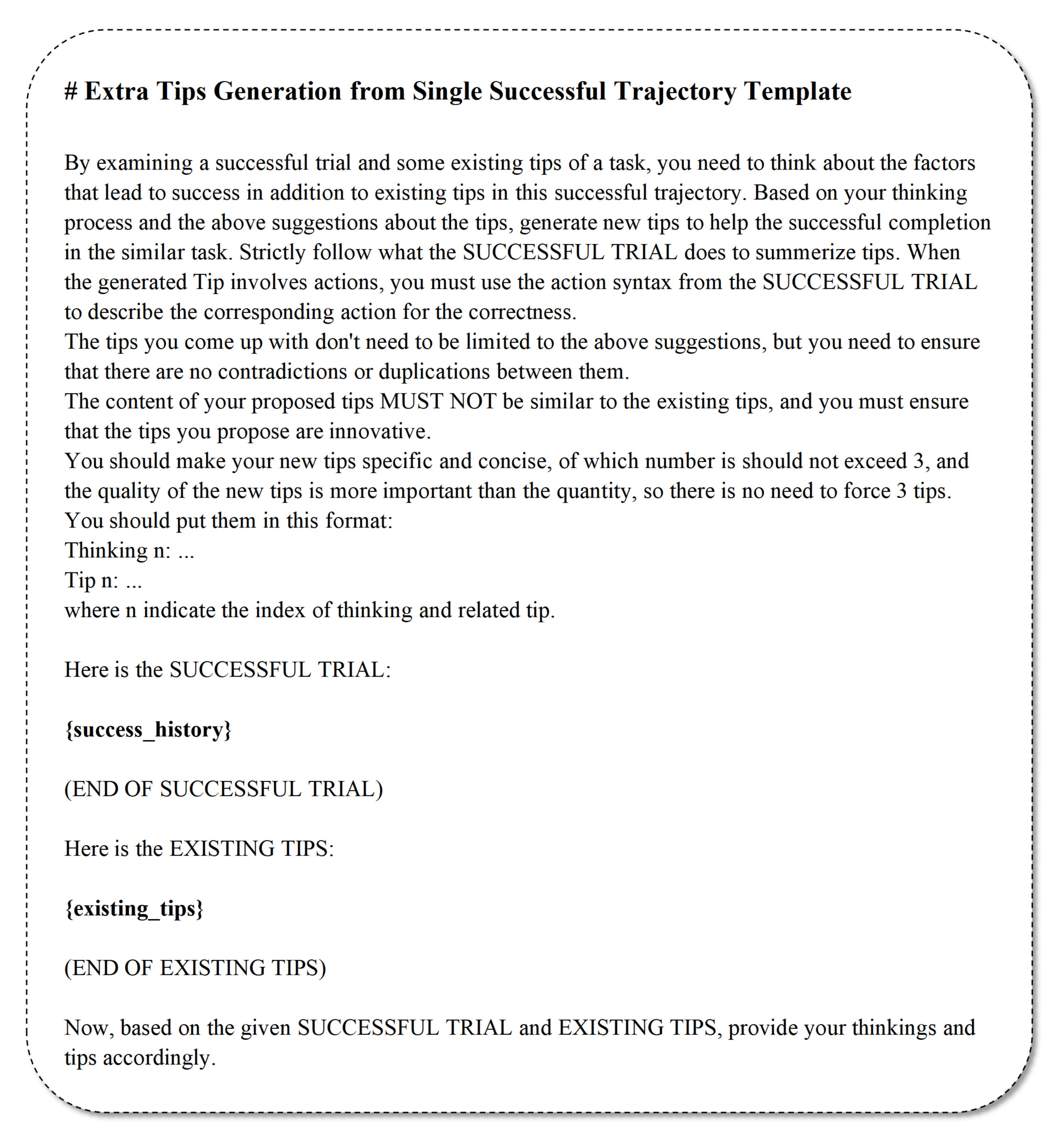} 
  \caption {The prompt template of extra tips generation from single success trajectory. The extraction is performed after generating tips from the compare trajectories as a supplement to get the factors leading to task success. }
  \label{fig:extra_tips_success}
\end{figure*}

\begin{figure*}[htbp]
  \centering
  \includegraphics[width=\linewidth]{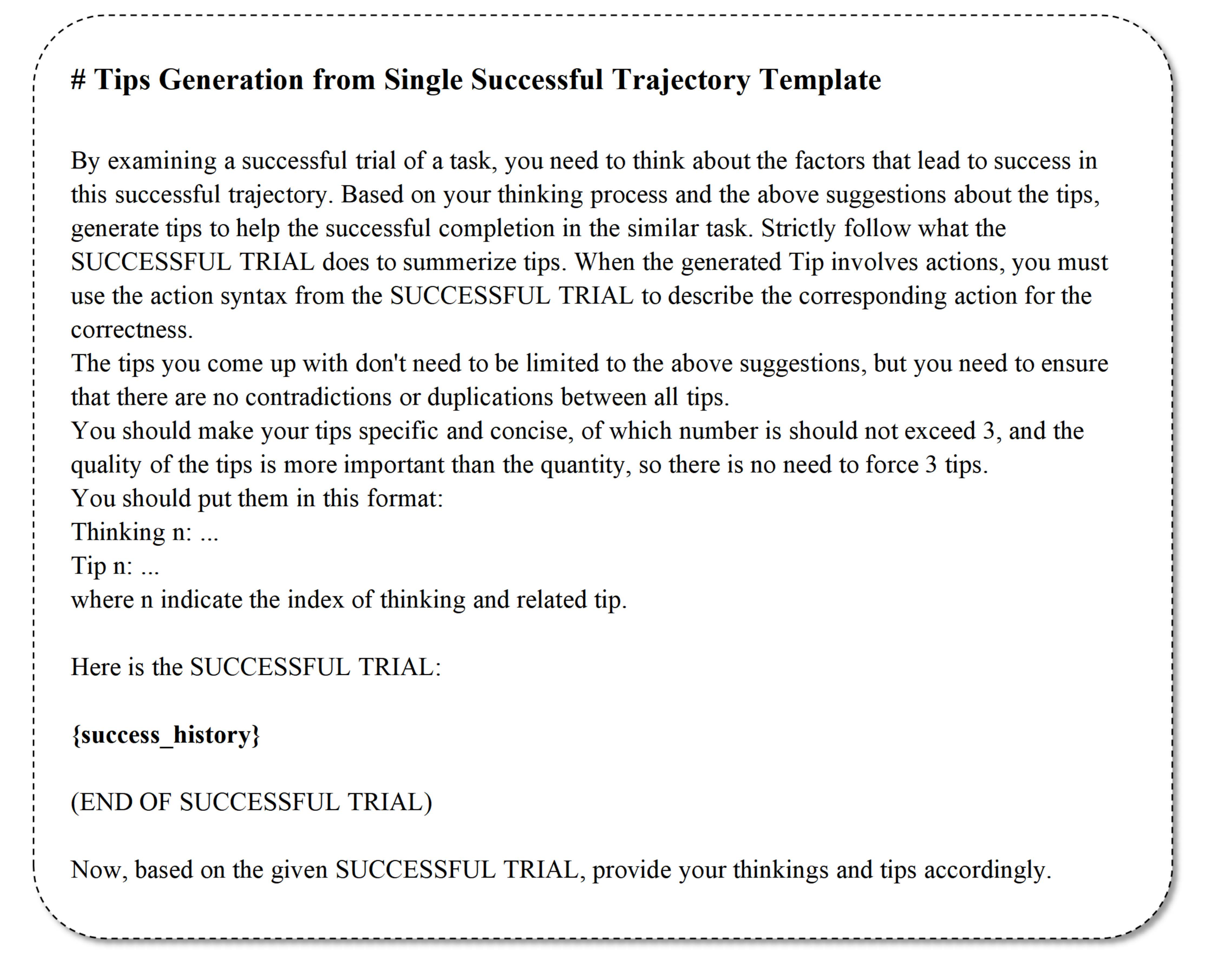} 
  \caption {The prompt template of tips generation from single success trajectory. The extraction targets the tasks which are successfully completed at the first try.For these tasks, there is only single success trajectory.}
  \label{fig:tips_success}
\end{figure*}

Figure \ref{fig:tips_compare}, \ref{fig:extra_tips_success} and \ref{fig:tips_success} detail our prompt template of experience-wise tips generation model in the ALFWorld, WebShop and ScienceWorld.

\subsection{Key Information Extraction Model}
\begin{figure*}[htbp]
  \centering
  \includegraphics[width=\linewidth]{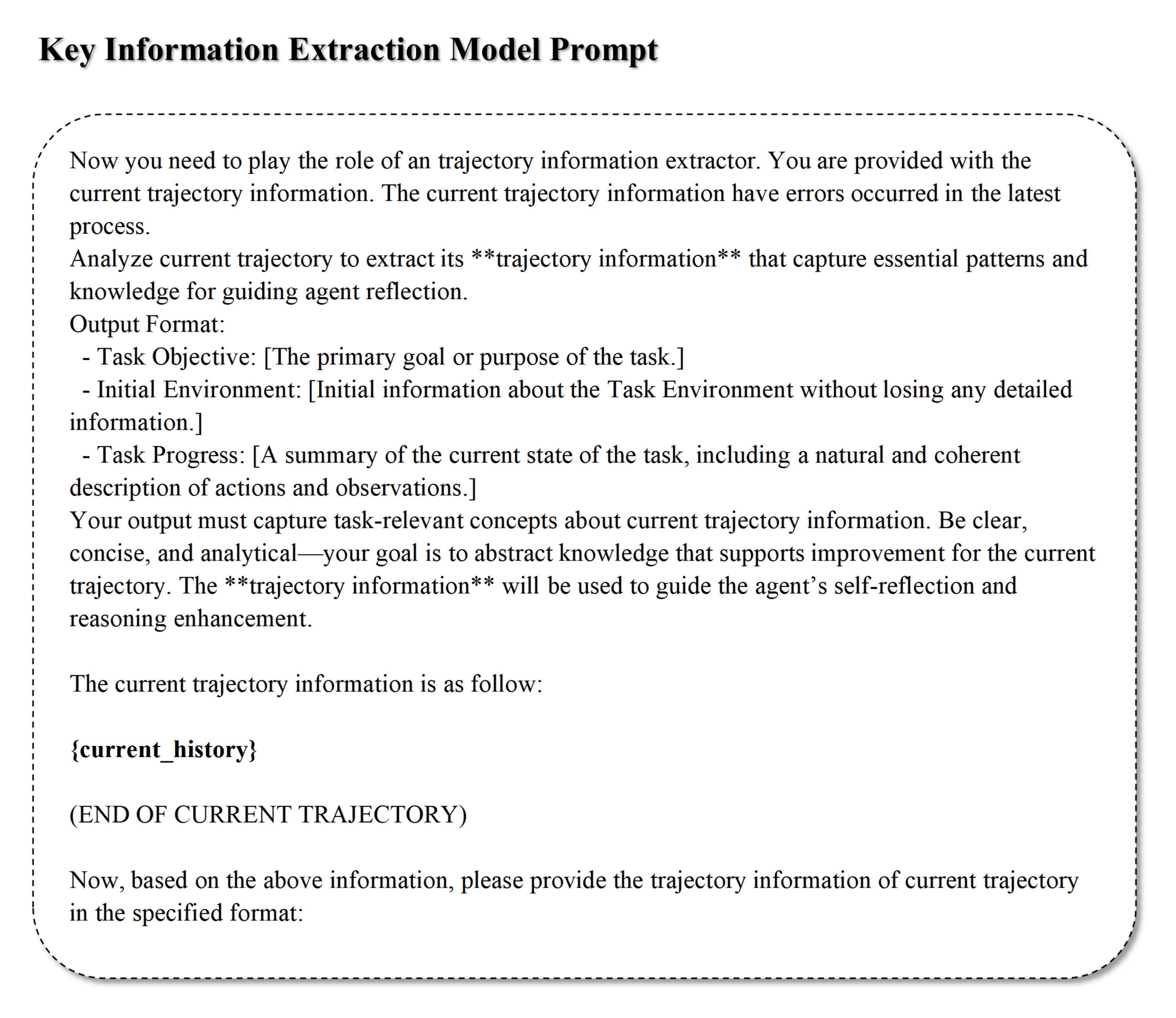} 
  \caption {The prompt template of key information extraction model in the ALFWorld, WebShop and ScienceWorld.}
  \label{fig:kie}
\end{figure*}

Figure \ref{fig:kie} shows our prompt template for key information extraction to perform a structured, fine-grained analysis of the current trajectory content.

\subsection{Key Information Reflection Model}
\begin{figure*}[htbp]
  \centering
  \includegraphics[width=\linewidth]{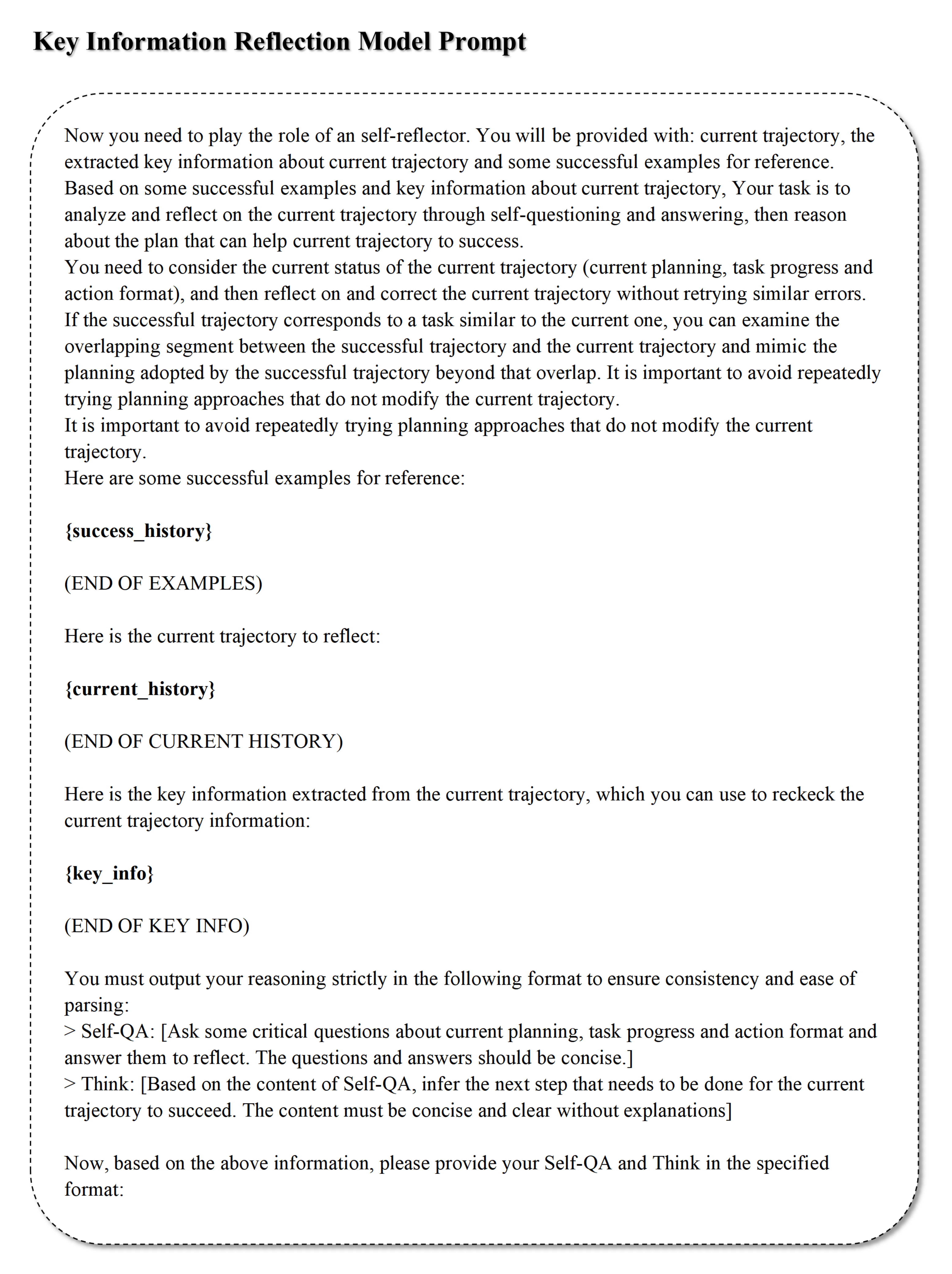} 
  \caption {The prompt template of key information reflection model in the ALFWorld, WebShop and ScienceWorld.}
  \label{fig:kir}
\end{figure*}

When specific abnormal observations happen during evaluation, we use the prompt with Figure \ref{fig:kir} for key information reflection.

\section{Example Trajectories}
\label{sec:examples}
\subsection{AlfWorld, \Ours{} \& ReAct}
\begin{figure*}[htbp]
  \centering
  \includegraphics[width=\linewidth]{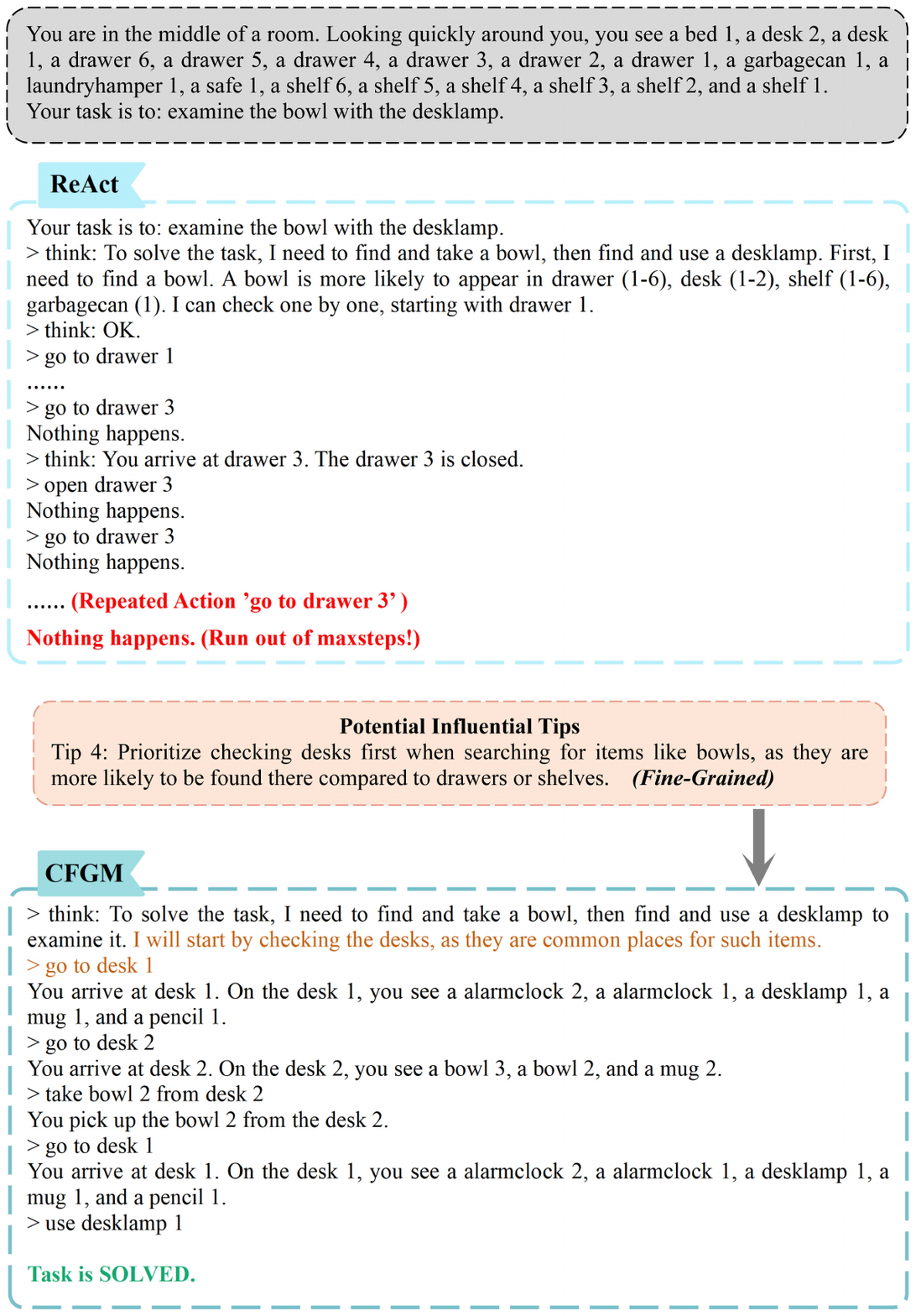} 
  \caption {Example Trajectories in the ALFWorld. The tip says, 'Prioritize checking the desk,' so the \Ours{} can quickly find the bowl instead of obsessing over drawer 3 and overlooking the more accessible solution on the desk—unlike ReAct, which ignored this efficient path.}
  \label{fig:compare_chosen4tips_alfworld}
\end{figure*}

\begin{figure*}[htbp]
  \centering
  \includegraphics[width=\linewidth]{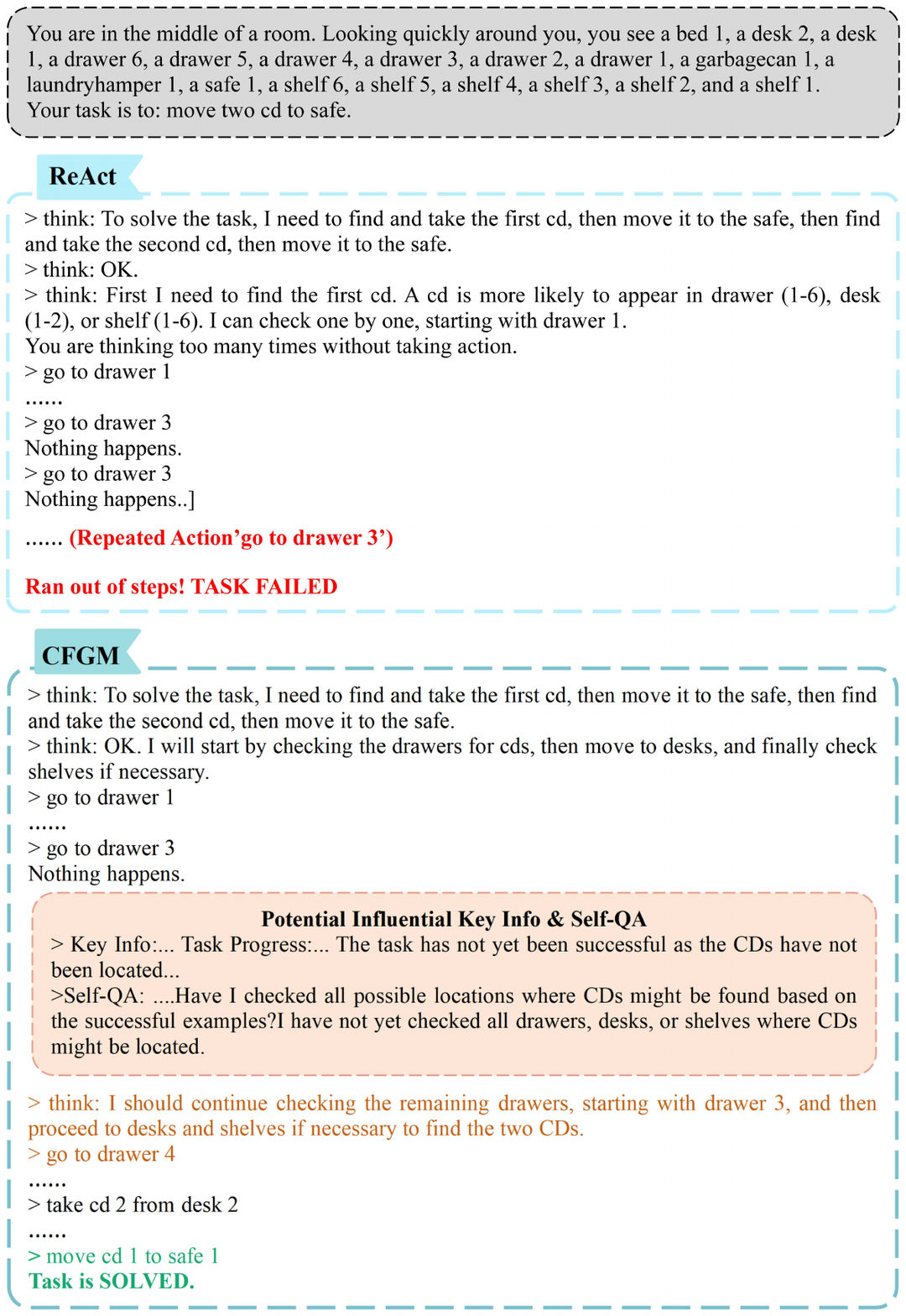} 
  \caption {Example Trajectories in the ALFWorld. When interacting with drawer 3 and the environment returns "Nothing happens", the \Ours{} realizes that it has not checked other locations by KIR. Instead of repeating the operation like ReAct, \Ours{} skips drawer 3 and continues to check other drawers.}
  \label{fig:compare_chosen4reflec_alfworld}
\end{figure*}

Figure \ref{fig:compare_chosen4tips_alfworld} and Figure \ref{fig:compare_chosen4reflec_alfworld} show the cases of tips and key information reflection in \Ours{} compared to ReAct, respectively, to help the agent avoid errors or recover from derailed planning in AlfWorld.
\subsection{WebShop, \Ours{} \& ReAct}
\begin{figure*}[htbp]
  \centering
  \includegraphics[width=\linewidth]{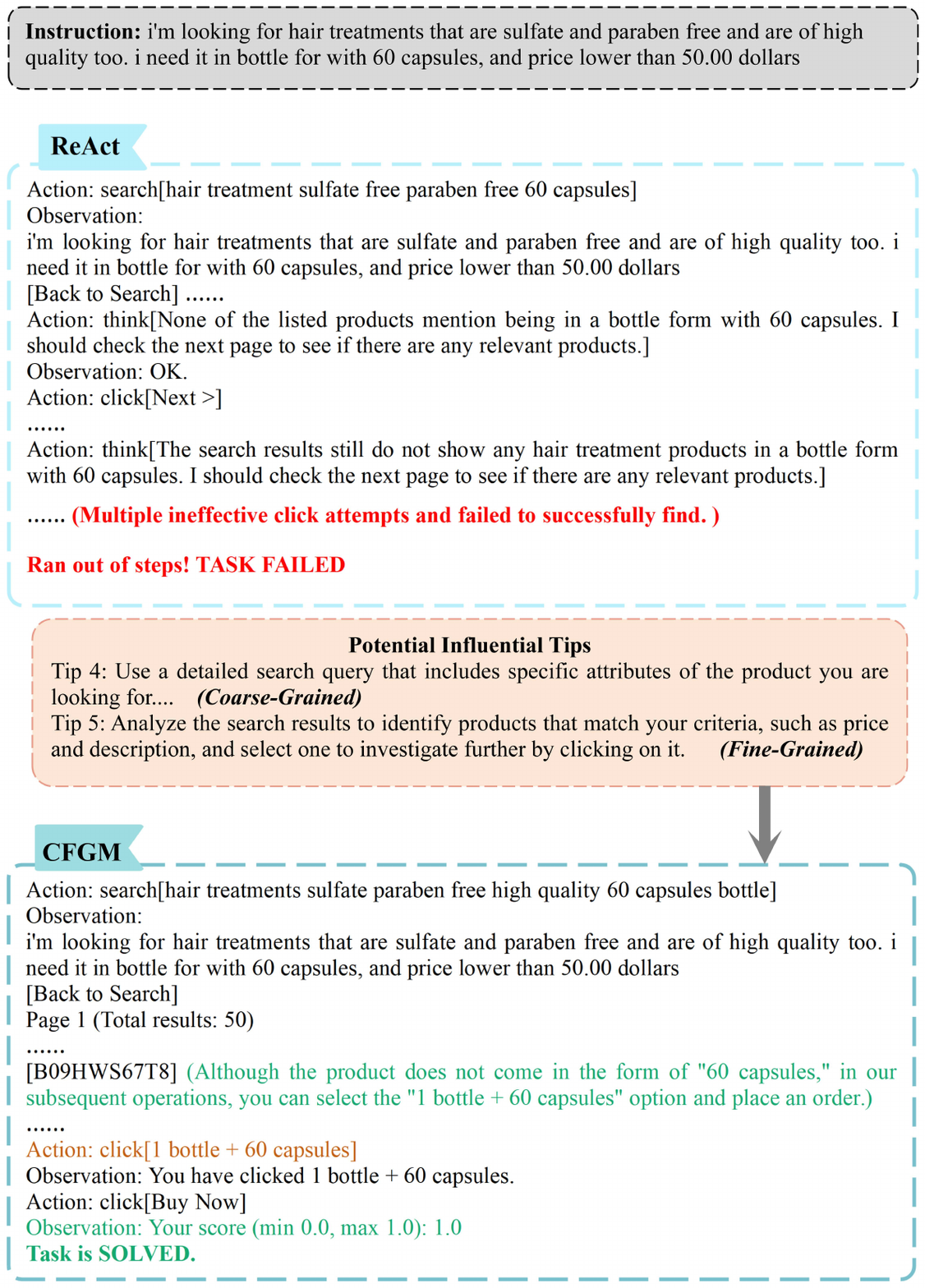} 
  \caption {Example Trajectories in the WebShop. The tips suggest using a detailed search query and investigating products by clicking on them, ensuring that \Ours{} focuses more precisely on target-matching products rather than relying on fuzzy searches like ReAct. Once a suitable product is found, \Ours{} proceeds to click on it and selects the '1 bottle, 60 capsules' option.}
  \label{fig:compare_chosen4tips_webshop}
\end{figure*}

\begin{figure*}[htbp]
  \centering
  \includegraphics[width=\linewidth]{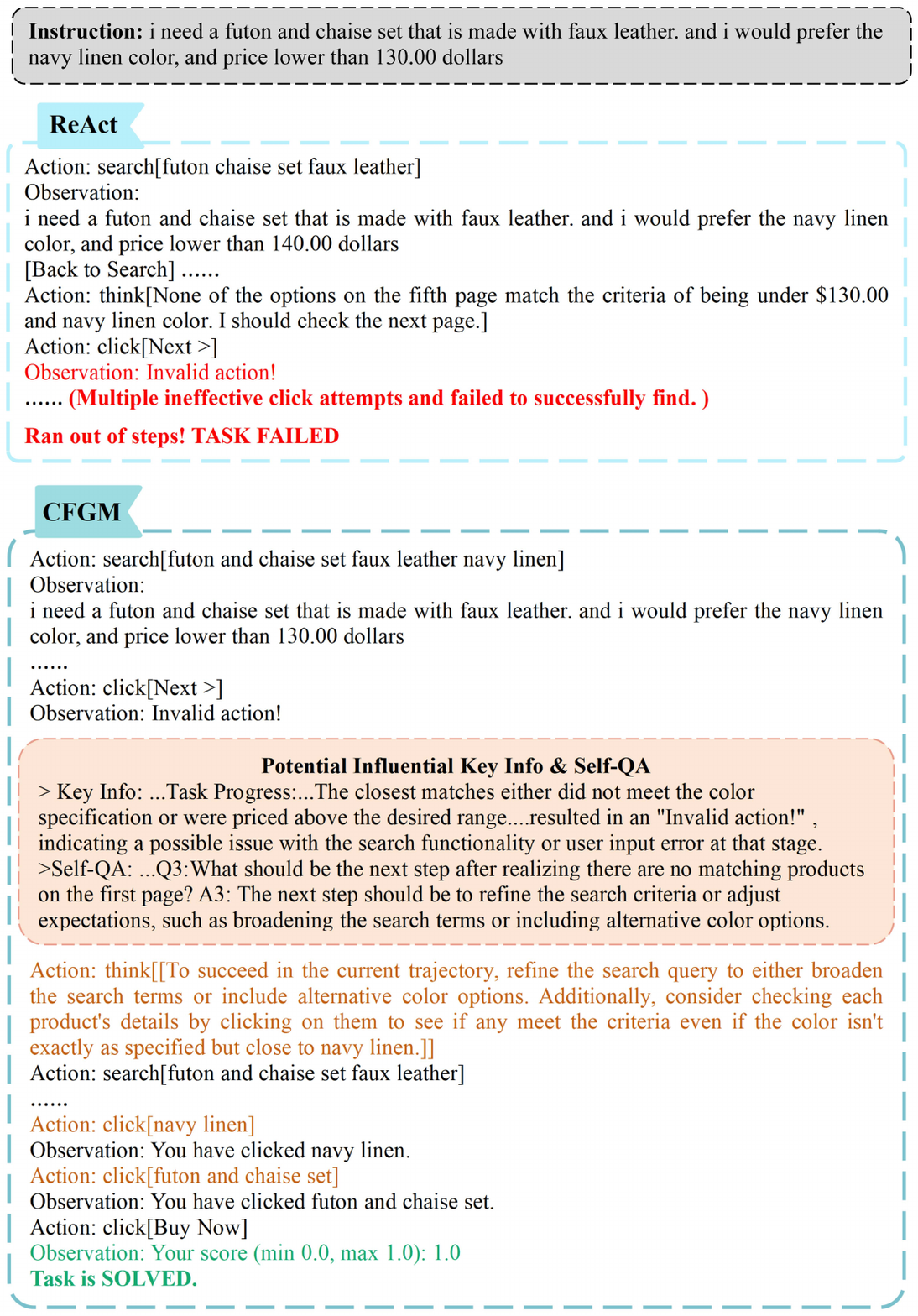} 
  \caption {Example Trajectories in the WebShop. Through KIR, \Ours{} recognized that the closest match in the current task progress failed to locate products with the desired color, prompting it to refine the search query by broadening the search terms. This approach avoided unnecessary multiple searches and page navigation—unlike ReAct, which exhausted its steps and ultimately failed.}
  \label{fig:compare_chosen4reflec_websop}
\end{figure*}

Figure \ref{fig:compare_chosen4tips_webshop} and Figure \ref{fig:compare_chosen4reflec_websop} show the cases of tips and key information reflection in \Ours{} compared to ReAct, respectively, to help the agent avoid errors or recover from derailed planning in WebShop.
\subsection{ScienceWorld, \Ours{} \& ReAct}
\begin{figure*}[htbp]
  \centering
  \includegraphics[width=\linewidth]{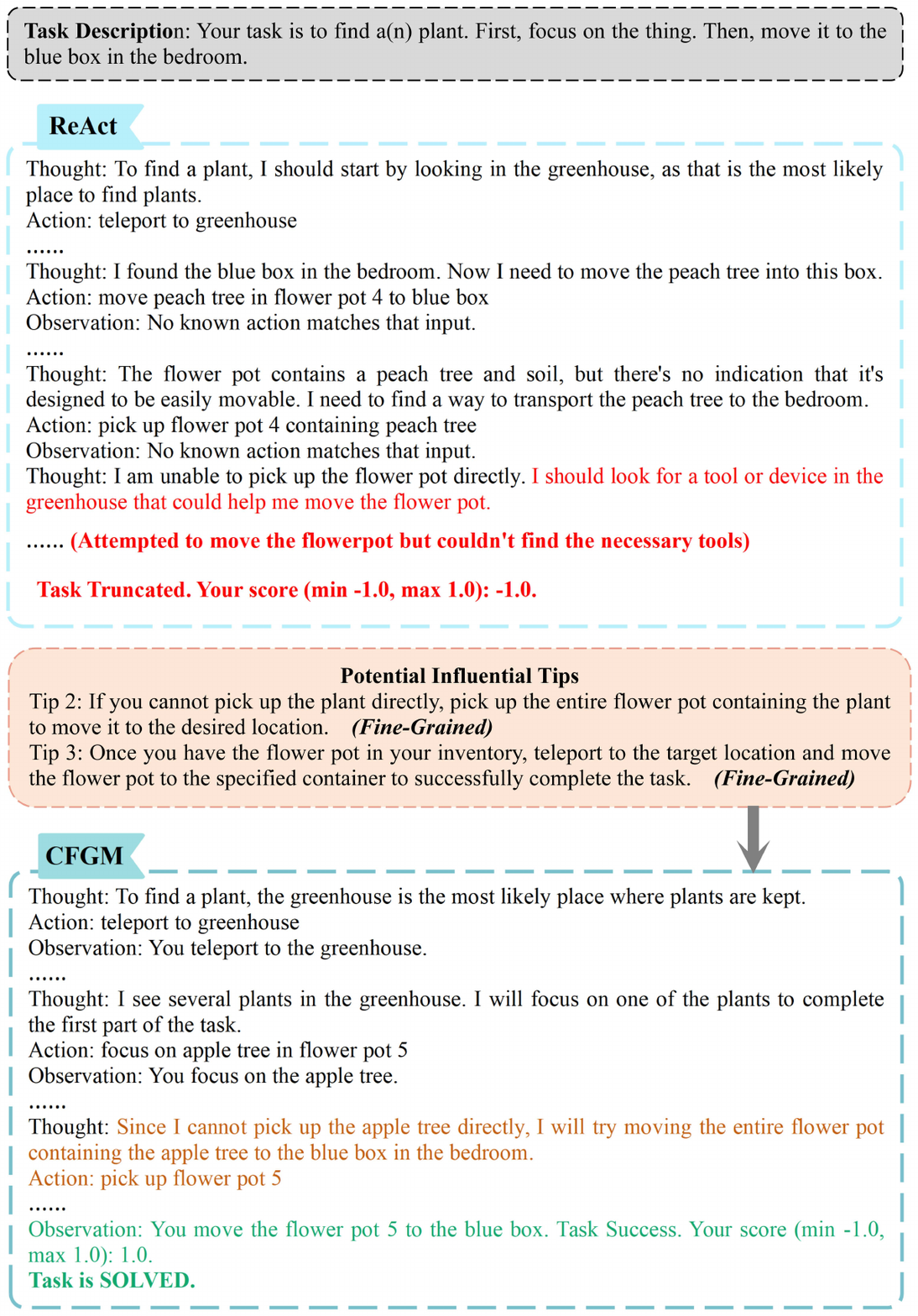} 
  \caption {Example Trajectories in the ScienceWorld. When \Ours{} realizes that it can't pick up the plant directly, tips reminds that it can move the whole pot and teleport to the target location immediately after moving it, avoiding the pointless search for a carrying tool in ReAct.}
  \label{fig:compare_chosen4tips_scienceworld}
\end{figure*}

\begin{figure*}[htbp]
  \centering
  \includegraphics[width=\linewidth]{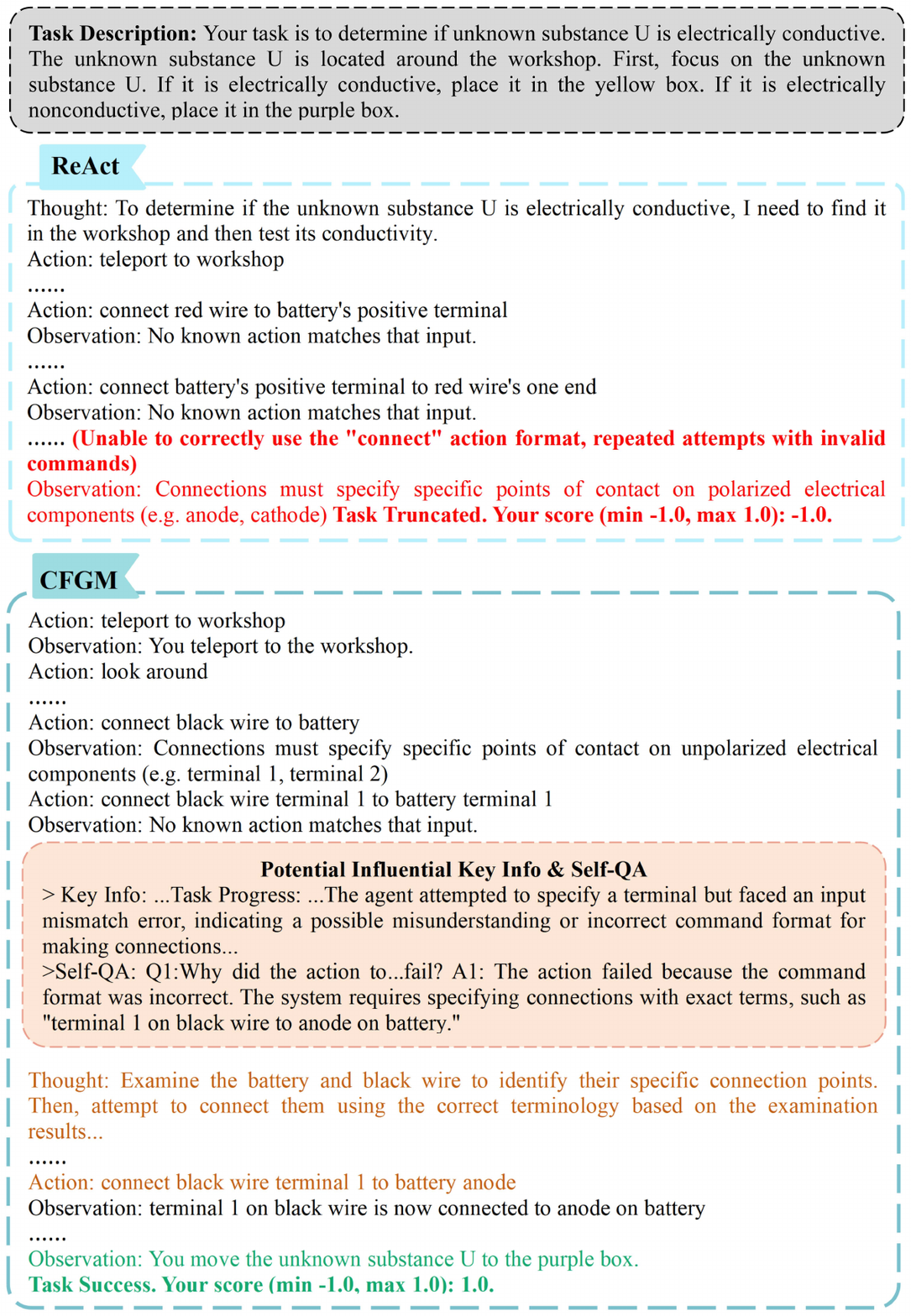} 
  \caption {Example Trajectories in the ScienceWorld. \Ours{} finds out that it is currently facing a mismatch error between the battery and the wire through the KIR which prompt a possible exact term to connect. Then \Ours{} successfully finds the correct way to connect by checking the connection points and avoids the invalid operation due to the formatting error in the ReAct.}
  \label{fig:compare_chosen4reflec_scienceworld}
\end{figure*}

Figure \ref{fig:compare_chosen4tips_scienceworld} and Figure \ref{fig:compare_chosen4reflec_scienceworld} show the tips and key information reflection in \Ours{}, respectively, to help the agent avoid errors or recover from derailed planning in ScienceWorld.



\end{document}